\documentclass{article}

\usepackage[margin=1in]{geometry}
\usepackage[english]{babel}
\usepackage{microtype}
\usepackage[utf8]{inputenc}
\usepackage{graphicx}
\usepackage{amsmath}
\usepackage{amsfonts}
\usepackage{graphicx}
\usepackage{amssymb,amsmath}
\usepackage{everypage}
\usepackage{xspace}
\usepackage{epsfig} 
\usepackage{tikz}
\usepackage{graphicx}
\usepackage{mathtools}
\usepackage{algpseudocode}
\usepackage{algorithm}
\usepackage{amsmath}
\usepackage{mathrsfs}
\usepackage{bm}
\usepackage{amsfonts}
\usepackage{color}
\usepackage{xcolor}
\usepackage{soul}
\usepackage{verbatim}
\usepackage{commath}
\usepackage{changepage}
\usepackage{multicol}
\usepackage{lipsum}    
\usepackage{lastpage} 
\usepackage{url} 
\usepackage{booktabs}
\usepackage[hang,font={small}]{caption}
\usepackage[font={small}]{subcaption}
\usepackage{rotating}
\usepackage{amsthm}
\usepackage{subcaption}
\usepackage{yhmath}
\usepackage{amsmath}
\usepackage{enumitem}
\usepackage{float}
\usepackage{amsmath,calc}
\usepackage{mathtools}
\usepackage{relsize}
\usepackage{wrapfig}
\usepackage{siunitx}
\usepackage[most]{tcolorbox}
\usepackage{fixltx2e} 
\usepackage{breqn}
\newcommand{\ignore}[1]{}

\usepackage{authblk}

\usepackage{lineno}

\usepackage{amsmath,empheq}
\usepackage{mathtools}
\usepackage[thinc]{esdiff}

\usepackage{color,soul}
\usepackage[nodisplayskipstretch]{setspace}
\newcommand{\R}{\mathbb{R}}

\usepackage{nomencl}
\makeglossary

\newtheorem{theorem}{Theorem}

\newcommand{\ts}{\textsuperscript}

\title{Learning to solve Bayesian inverse problems: An amortized variational inference approach \textcolor{black}{ using Gaussian and Flow guides}}
\author[1]{Sharmila Karumuri\thanks{Corresponding author: skarumur@purdue.edu}}
\author[1]{Ilias Bilionis}
\affil[1]{School of Mechanical Engineering, Purdue University, West Lafayette, IN, USA}

\begin{document}

\maketitle


Inverse problems, i.e., estimating parameters of physical models from experimental data, are ubiquitous in science and engineering.
The Bayesian formulation is the gold standard because it alleviates ill-posedness issues and quantifies epistemic uncertainty.
Since analytical posteriors are not typically available, one resorts to Markov chain Monte Carlo sampling or approximate variational inference.
However, inference needs to be rerun from scratch for each new set of data.
This drawback limits the applicability of the Bayesian formulation to real-time settings, e.g., health monitoring of engineered systems, and medical diagnosis.
The objective of this paper is to develop a methodology that enables real-time inference by learning the Bayesian inverse map, i.e., the map from data to posteriors.
Our approach is as follows.
\textcolor{black}{
We parameterize the posterior distribution as a function of data. 
This work outlines two distinct approaches to do this. 
The first method involves parameterizing the posterior using an amortized full-rank Gaussian guide, implemented through neural networks. 
The second method utilizes a Conditional Normalizing Flow guide, employing conditional invertible neural networks for cases where the target posterior is arbitrarily complex. 
In both approaches,  we learn the network parameters by amortized variational inference which involves maximizing the expectation of evidence lower bound over all possible datasets compatible with the model.}
We demonstrate our approach by solving a set of benchmark problems from science and engineering.
Our results show that the posterior estimates of our approach are in agreement with the corresponding ground truth obtained by Markov chain Monte Carlo.
Once trained, our approach provides the posterior distribution for a given observation just at the cost of a forward pass of the neural network.



\section{Introduction}
\noindent

In scientific and engineering applications, we are often interested in identifying the unknown parameters of a physical model from observable quantities.
These problems are called inverse or model calibration problems~\cite{aster2018parameter}.
For instance, in reservoir engineering, it is pivotal to infer the permeability field of the subsurface from geophysical field measurements~\cite{oliver2008inverse}. 
Other examples of inverse problems include remote sensing~\cite{haario2004markov}, climate modeling~\cite{bilionis2015crop}, medical imaging~\cite{aguilo2010inverse}, subsurface hydrology and geology~\cite{hill2006effective}, ocean dynamics~\cite{herbei2008gyres}, seismic inversion~\cite{russell1988introduction}, and many more.

Inverse problems are hard to solve.
First, the observed data typically contain measurement noise which has to be filtered out.
Second, inverse problems may be ill-posed, i.e., many different sets of parameters could result in the same observations.
Third, forward models are usually computationally expensive with simulation times ranging from a few minutes to days.

Bayesian inference is the gold standard for posing inverse problems~\cite{tarantola2005inverse, jaynes2003probability}.
In the Bayesian paradigm, one encodes their knowledge about the parameters using prior probabilities, and models the measurement process using a likelihood function which connects the physical model to the data.
The solution of the inverse problem is the posterior probability dictated by Bayes' rule~\cite{calvetti2018inverse}.
The analytical form of the posterior is not always available, except for very few simple cases.

The crudest way to summarize the Bayesian solution is via a point estimate of the parameters, typically obtained by maximizing the posterior probability density (MAP estimate).
This approach is used in seismic inversion~\cite{fichtner2010full} and numerical weather prediction models~\cite{navon2009data}.
MAP estimates are acceptable only when the posterior has a unique maximum and is sharply peaked.

The Laplace method~\cite{ghosh2006introduction} approximations the posterior using a multivariate Gaussian with a mean specified by the MAP estimate and a covariance matrix given by the negative inverse Hessian of the logarithm of the posterior.
This approximation is capable of quantifying some of the epistemic uncertainty, albeit it is acceptable only in the cases where the posterior has a unique maximum and is shaped like a Gaussian.

More sophisticated approaches involve exploring the posterior by sampling through Markov chain Monte Carlo (MCMC)~\cite{bishop:2006:PRML, jones2021markov, laloy2012high} sampling methods. 
MCMC generates a sequence of samples from a proposal distribution which are accepted or rejected according to an acceptance ratio.
The final samples form a Markov chain that is ergodic with respect to the desired posterior.
MCMC requires repeated evaluations of the underlying physical model, which raises the computing overhead.
This computational cost can be overcome by replacing the physical model with a computationally inexpensive surrogate.
Surrogates are built, for example, using Gaussian process regression (GPR)~\cite{martin2005use, bilionis2012multi, bilionis2013multi, chen2015uncertainty, tripathy2016gaussian, bilionis2013solution}, Polynomial chaos expansion (PCE)~\cite{najm2009uncertainty, eldred2009comparison, xiu2002wiener}, Deep neural networks (DNNs)~\cite{karumuri2020simulator, zhu2018bayesian,tripathy2018deep}.
Note that surrogate models introduce additional epistemic uncertainty, an issue that can be addressed using the theory developed in~\cite{bilionis2013solution}.
MCMC is not without its issues.
For instance, as the number of parameters increases; the generated Markov chain may take impractically long times to converge~\cite{van2018simple, papamarkou2019challenges}.
\textcolor{black}{In ~\cite{wang2004bayesian}, an MCMC-based Bayesian approach was applied to solve an inverse heat conduction problem.}
\textcolor{black}{Authors of~\cite{lieberman2010parameter, cui2016scalable, chen2016sparse} proposed projection-based reduced-order models for accelerating
the computation of Bayesian inverse problems using MCMC.}

Variational inference (VI)~\cite{blei2017variational, ranganath2014black, ganguly2023amortized} offers a compromise between computational efficiency and accuracy.
The idea is to pose the posterior learning problem as an optimization problem over a family of tractable probability distributions.
The optimization objective is usually the information loss between the true and the approximate posterior.
Common choices are the Kullback–Leibler (KL) divergence~\cite{tsilifis2014variational} and Maximum mean discrepancy (MMD)~\cite{pomponi2021bayesian}.
The parameters of the approximated posterior are referred to as variational parameters and the corresponding approximated posterior as the variational distribution or as the guide.
In~\cite{tsilifis2016computationally}, the authors applied the VI formulation to approximate the posterior of an inverse problem by a family of a mixture of Gaussians with applications to catalysis and contamination source identification.
Similarly,~\cite{jin2012variational} and~\cite{franck2016sparse} applied these techniques for parameter inference in heat conduction and elastography.


A major drawback of the presented methodologies is that they need to be rerun for each new set of observations.
As a result, it is not always feasible to apply these methods to settings that require a real-time response.
Overcoming this limitation has a potential impact on many applications, e.g., medical imaging~\cite{lipkova2019personalized, kershaw1999application}, structural health monitoring~\cite{sajedi2021uncertainty, wu2020data}, geology~\cite{de2016structural}.
The goal of our paper is to address this drawback.
More specifically, our objective is to develop a methodology that enables real-time inference by learning a generalized model that outputs the posterior for any observed data that is compatible with the physical process.
We refer to this function from data to parameters as the ``Bayesian inverse map.''

The idea of learning inverse maps has been explored in previous works.
The authors of~\cite{ardizzone2018analyzing}, employed an invertible neural network to learn both the forward map (from parameters to data) and the inverse map.
However, this work is incomplete in the sense that they assumed that their data was noise-free.
In\textcolor{black}{~\cite{padmanabha2021solving, radev2020bayesflow, radev2023bayesflow, kruse2021hint}, the authors proposed to learn inverse maps by parameterizing the posterior as a deep-generative conditional flow model~\cite{ardizzone2019guided}, where a sequence of invertible transformations to a base conditional distribution models the posterior.
These invertible transformations are parameterized functions represented by neural networks.
They trained the parameters of these networks on pairs of parameter-observation data by maximizing the conditional likelihood or by minimizing the KL divergence between true and approximate posterior for all possible
datasets.
It can be shown that this approach is a posterior mean-seeking approach and not a mode-seeking approach.
The latter point creates problems when one tries to apply the method to ill-posed inverse problems.}
Our method addresses the same problems.

Next, the authors of~\cite{kaltenbach2023semi}, introduced an invertible DeepONet architecture to learn inverse maps, however, they estimate the posterior through a semi-analytic approach.
\textcolor{black}{In~\cite{cui2023scalable},  authors explored the transport map~\cite{el2012bayesian, marzouk2016sampling} idea to solve Bayesian inverse problems in real-time.
They proposed an offline-online approach, wherein the joint law of parameter and observable random variables is learned via tensor-train decomposition~\cite{bigoni2016spectral, gorodetsky2019continuous} based density approximation during the offline phase.
In the online phase, the resulting conditional map facilitates real-time computation of the posterior distribution for given observed data. 
The authors adopted a function approximation perspective for learning these transport maps by minimizing the Hellinger distance between the approximate and the true posterior densities.
However, this method encounters challenges when we have a large number of parameters with concentrated density functions and complex nonlinear interactions.}

We represent the Bayesian inverse map using amortized variational distributions~\cite{gershman2014amortized, zhang2018advances, margossian2023amortized}. 
Amortized variational distributions are guides with parameters that are functions of the observed data.
These functions accept observations as inputs and output the parameters of the guide representing the Bayesian solution to the inverse problem.
We represent these functions using a neural network, called the amortization network.
We identify the parameters of the amortization network by minimizing the expectation (over all datasets supported by the physical model) of the Kullback-Leibler divergence between the guide and the true posterior.
We call our approach amortized variational inference (AVI).
We prove theoretically that, under certain assumptions, optimizing the proposed objective function is equivalent to solving all possible VI problems in one shot.

We also derive a stochastic optimization algorithm that enables the practical implementation of our scheme.
The problem is very computationally demanding, but the cost is ``amortized'' when the solution is repeatedly used.
Most importantly, the Bayesian inverse map can be queried at the cost of a single forward amortization network pass and, thus, it is suitable for real-time applications.
Note that AVI is more restricted than free VI, a phenomenon called the amortization gap~\cite{cremer2018inference, choi2019meta}. 
Of course, as the amortization network capacity goes to infinity, the amortization gap disappears.
In practice, one has to balance the amortization network capacity with the available computational resources for identifying its parameters.


The rest of the paper is structured as follows.
In Sec.~\ref{sec:methodology}, we outline our methodology by first discussing the mathematical notation used throughout the paper. In Sec.~\ref{subsec:prob_def}, we describe the problem we intend to solve. We then describe in detail the variational formulation to learn the inverse map using amortized posteriors in Sec.~\ref{subsec:variational_formulation}.
\textcolor{black}{In Sec.~\ref{subsec:amortized_net}, we move on to the discussion of the choice of the amortized posteriors used in this work and their representation using neural networks.}
Finally, in Sec.~\ref{subsec:optimization}, we discuss the stochastic optimization of the variational loss.
In Sec.~\ref{sec:results}, we discuss the metrics used for evaluating the performance of our approach and then demonstrate the methodology on a series of examples.
In Sec.~\ref{sec:conclusion} we present our concluding remarks.

\section{Methodology}
\label{sec:methodology}
\noindent

We start with a discussion of the mathematical notation we follow regarding random variables and their expectations.
We use uppercase letters to indicate random variables and lowercase letters to indicate the values of these random variables.
We assume that all the random variables we are working with have probability densities.
If the probability density of a random variable $X$ is not explicitly specified, then we denote it by $p(x)$.
In this regard, we follow the common practice in probabilistic machine learning of ``overloading'' the symbol $p$.
In particular, when we encounter the symbols ``$p(x)$'' then we understand that it refers to ``the probability density function of the random variable $\text{upper}(x) = X$ evaluated at $x$.''

Now if $g$ is a function of $x$, the expectation of $g(X)$ is:
$$
\mathbb{E}[g(X)] = \int g(x)p(x)dx.
$$
Sometimes we want to take the expectation of $X$ not with respect to $p(x)$ but with respect to another distribution, say $q$.
We denote this expectation by:
$$
\mathbb{E}_{X\sim q}[g(X)] = \int g(x) q(x) dx.
$$
When there is no ambiguity, we may simply write $\mathbb{E}_q$ instead of $\mathbb{E}_{X\sim q}$.
We define the (differential) entropy of the probability density $q(x)$ by:
$$
\mathbb{H}[q(X)] := -\mathbb{E}_{q}[\log q(X)].
$$
Finally, we denote by $\mathcal{N}(x|\mu,\Sigma)$ the probability density of a multivariate Gaussian with mean $\mu$ and covariance matrix $\Sigma$ evaluated at $x$.

\subsection{Problem definition and motivation}
\label{subsec:prob_def}
\noindent

Suppose that we have a physical problem with unknown parameters as $\xi$, a vector in $\R^{d}$.
The physical model connects the parameter $\xi$ to some quantities of interest. 
We refer to this map as the ``forward model.''
The forward model is a function, $f$, from $\R^d$ to $\R^m$.
The evaluation of the forward model at a given parameter vector is $f(\xi)$.
We denote the experimental observations by $y$, also a vector in $\R^m$.
The goal in inverse problems is to find the parameters $\xi$ from the data $y$, i.e., to invert the forward model.

Note that the data differ from the model prediction due to a variety of reasons, e.g., measurement noise, model discrepancy errors, errors due to the discretization of the physical equations, and numerical errors.
We work under the simplifying assumption that only measurement uncertainty is present.
For concreteness, let us assume that the data are generated by adding zero-mean Gaussian noise to the forward model.
The likelihood function, which connects parameters to data, is:
\begin{equation*}
p(y|\xi) = \mathcal{N}(y|f(\xi), \gamma^2I),
\end{equation*}
where the mean is the model prediction, $I$ is the unit matrix, and $\gamma^2$ is a parameter that controls the measurement noise.

In the Bayesian formulation of inverse problems, one starts by describing their state of knowledge about the parameters $\xi$ using probabilities.
Let $\Xi$ be the random variable encoding this prior knowledge and $p(\xi)$ the corresponding prior probability density function.
After observing the data, we wish to update our state of knowledge about the parameters.
Following the Bayesian paradigm, our posterior state of knowledge is captured by:
\begin{equation*}
    p(\xi|y) = \frac{p(y|\xi) p(\xi)}{p(y)}.
\end{equation*}
The normalizing constant $p(y) = \int p(y|\xi) p(\xi) \, d{\xi}$, is known as the evidence.

The posterior is, typically, not analytically available.
In VI, one approximates it within a tractable distribution family $q_{\lambda}(\xi)$.
We call $q_{\lambda}(\xi)$ the guide and we refer to $\lambda$ as the variational parameters.
One identifies the variational parameters by minimizing the Kullback–Leibler (KL) divergence between the guide and the posterior.
KL minimization is equivalent to maximizing a related quantity, the Evidence Lower BOund (ELBO)~\cite{bishop2006pattern,kucukelbir2016automatic}.

Equipped with this notation, we can write the mathematical equation for the ELBO.
It consists of two parts.
The first part, which promotes data fitting, is the expectation over the guide of the logarithm of the joint probability density of parameters and data.
The second part, which serves a regularization purpose, is the entropy of the guide.
If we use $p(\xi, y) = p(y|\xi)p(\xi)$ to denote the joint probability density of parameters and data, then the mathematical form of the ELBO is:
\begin{equation}
\label{eqn:ELBO}
\text{ELBO}(\lambda;y) := \mathbb{E}_{q_{\lambda}} \left[\log p(\Xi,y)\right] + \mathbb{H}\left[q_{\lambda}(\Xi)\right],
\end{equation}
\textcolor{black}{where, $\mathbb{H}\left[q_{\lambda}(\Xi)\right]= \mathbb{E}_{q_{\lambda}}[-\log (q_\lambda(\Xi))]$ is the  differential entropy of the guide.}

\subsection{Variational formulation of the problem of finding the inverse map}
\label{subsec:variational_formulation}
\noindent

The main drawback of VI is that it requires solving the variational problem for each new data.
This shortcoming inhibits the application of VI to real-time inference settings. 
Our goal is to learn the inverse map, i.e., the map from data to posteriors.
To this end, we rely on two pillars.
First, we use amortization to represent the map from data to optimal variational parameters.
Second, we formulate a new variational problem whose solution is equivalent to doing VI for all datasets compatible with the model.

The idea in amortization is to make the optimal variational parameters a function of the data, i.e., $\lambda = \lambda(y)$.
So, for new data $y$, the posterior is approximated by the guide $q_{\lambda(y)}(\xi)$.
We refer to $\lambda$ as the amortization function.
For concreteness, assume that there are $n$ variational parameters so that $\lambda$ is a function from $\R^m$ to $\R^n$.
Let $\lambda_i$ denote the $i$-th component of $\lambda$.
We define the space of admissible amortization functions, $\mathcal{A}$, to be the set of Lebesgue-measurable functions $\lambda$ from $\R^m$ to $\R^n$ with finite $L^2$ norm:
$$
\parallel\lambda\parallel^2 := \mathbb{E}\left[\sum_{i=1}^n\lambda_i^2(Y)\right] < \infty.
$$
The space of admissible amortization functions $\mathcal{A}$ is a Banach space.

Note that the expectation above is over the data random variable $Y$ which is assumed to follow the data density $p(y)$, as predicted by our model.
Using the sum rule, the probability density function of $Y$ is:
$$
p(y) = \int p(\xi, y) d\xi = \int p(y|\xi)p(\xi)d\xi.
$$
In other words, one can sample $Y$ by sampling parameters from the prior, evaluating the forward model, and then sampling data from the likelihood.

We propose to learn the amortization function by maximizing the expectation of the ELBO, Eq.~(\ref{eqn:ELBO}), over all admissible amortization functions:
\begin{equation}
\text{AELBO}[\lambda] = \mathbb{E}\Big[\text{ELBO}(\lambda(Y);Y)\Big].
\label{eqn:Amortized ELBO}
\end{equation}
This expectation is well defined whenever the ELBO is continuous (the composition of a continuous function with a Lebesgue-measurable function is Lebesgue-measurable).
We refer to this quantity as the amortized ELBO (or AELBO).

Next, we prove two \textcolor{black}{theorems} that provide some intuition as to why the AELBO is a good choice for learning the amortization function.
The first \textcolor{black}{theorem} claims that the AELBO is bounded above by minus the differential entropy of the data density.
Observe that this statement does not necessarily mean that the AELBO has an attainable maximum.
But the statement guarantees that a maximization algorithm will not result in perpetual AELBO increase.

\begin{theorem}
If the differential entropy of the data density, $\mathbb{H}[p(Y)]$, is finite, then the amortized ELBO is bounded above by -$\mathbb{H}[p(Y)]$ for all admissible amortization functions.
\end{theorem}
\begin{proof}
Let $\lambda$ be an admissible amortization function.
The ELBO is bounded above by the log evidence~\cite{tsilifis2016computationally}, i.e.,
$$
\text{ELBO}(\lambda(Y);Y) \le \log p(Y).
$$
Taking the expectation of both sides with respect to the data density yields:
$$
\text{AELBO}[\lambda] \le \mathbb{E}[\log p(Y)] = -\mathbb{H}[p(Y)].
$$
\end{proof}

It is also worth noting that one can construct pathological probability densities whose differential entropy is minus infinity.
For such cases, our argument breaks down.
It is also possible to construct probability densities with infinite differential entropy.
For such data densities, the steps in the proof show that the AELBO is minus infinity and, thus, meaningless.
In what follows, we are assuming that the data density has a finite differential entropy.
We refer the interested reader to the work of~\cite{6089568} for sufficient conditions under which this assumption is true.

Let $\lambda$ and $\zeta$ be admissible amortization functions.
The first variation of the AELBO with respect to $\lambda$ in the direction of $\zeta$ is defined by:
$$
\delta\text{AELBO}[\lambda,\zeta] := \left.\frac{d}{d\epsilon}\right|_{\epsilon=0}\text{AELBO}[\lambda + \epsilon\zeta].
$$
The second variation of the AELBO at $\lambda$ in the direction of $\zeta$ is:
$$
\delta^2\text{AELBO}[\lambda,\zeta] := \left.\frac{d^2}{d\epsilon^2}\right|_{\epsilon=0}\text{AELBO}[\lambda + \epsilon\zeta].
$$
The necessary and sufficient conditions for an admissible amortization function $\lambda$ to be a \textcolor{black}{local} maximum of AELBO is that the first variation is zero and the second variation is strongly negative for all directions $\zeta$ in $\mathcal{A}$~\cite{gelfand2000calculus}, i.e.,
$$
\delta\text{AELBO}[\lambda,\zeta] = 0,
$$
and
$$
\delta^2\text{AELBO}[\lambda, \zeta] < - \kappa\parallel \zeta\parallel^2,
$$
for some $\kappa > 0$.
Similarly, if a variational parameter $\lambda(y)$ maximizes the ELBO \textcolor{black}{locally,} then the gradient of the ELBO is zero at $\lambda(y)$ and the Hessian of the ELBO is negative definite.
The next \textcolor{black}{theorem} guarantees that \textcolor{black}{local} maxima of the AELBO yield \textcolor{black}{local} maxima of the ELBO.
Note that there are underlying technical smoothness assumptions which we do not explicitly state.
The reader should assume that the functions involved are as smooth as necessary for the steps of the proof to be valid.
\textcolor{black}{Also, the other assumption is that there is no amortization gap, i.e., the admissible amortization function space} $\color{black} \mathcal{A}$ \textcolor{black}{is extensive enough and encompasses all the local maxima of the ELBO,} $\color{black} \lambda(y).$

\begin{theorem}
    \textcolor{black}{Assume that $\text{ELBO}(\lambda,y)$ is a twice continuously differentiable function of the parameters $\lambda$.}
    If an admissible amortization function, $\lambda$, is a \textcolor{black}{local} maximum of the amortized ELBO, then the variational parameters $\lambda(y)$ form a \textcolor{black}{local} maximum of the ELBO for all data $y$ supported by the data density.
\end{theorem}
\begin{proof}
To keep the notation as simple as possible, define the function $g$ from $\R^n\times \R^m$ to $\R$ by:
$$
g(\lambda,y) = \text{ELBO}(\lambda,y).
$$
The AELBO is the functional from $\mathcal{A}$ to $\R$:
$$
\text{AELBO}[\lambda] = \mathbb{E}[g(\lambda(Y),Y)],
$$
where the expectation is with respect to the random vector $Y$ which follows the data density.

Let $\lambda$ be an admissible amortization function that \textcolor{black}{locally} maximizes the AELBO.
We will show that $\lambda(y)$ \textcolor{black}{locally} maximizes the ELBO for all $y$ in the support of the data density.
The first variation of $\text{AELBO}[\lambda]$ in an arbitrary direction $\zeta$ must be zero.
Using the chain rule, we get:
\begin{equation}
\label{eqn:first-variation}
0 = \delta\text{AELBO}[\lambda,\zeta] = \mathbb{E}\left[\sum_{i=1}^n\frac{\partial g(\lambda(Y),Y)}{\partial \lambda_i}\zeta_i(Y)\right].
\end{equation}

Now for any $j=1,\dots,n$ and any $y$ in the support of the data density, pick a $\zeta$ whose components are the product of the following carefully chosen Kronecker and Dirac deltas:
$$
\zeta_i(Y) = \delta_{ij}\delta(Y-y).
$$
Plugging in Eq.~(\ref{eqn:first-variation}) yields:
$$
\frac{\partial g(\lambda(y),y)}{\partial \lambda_j} = 0.
$$
This is the necessary condition for $\lambda(y)$ to be a \textcolor{black}{local} maximum of the ELBO.

Since $\lambda$ is a \textcolor{black}{local} maximum of the AELBO, the second variation is strictly negative.
This means that there exists a positive $\kappa$ such that for all $\zeta$:
$$
\delta^2 \text{AELBO}[\lambda,\zeta] < -\kappa \parallel \zeta\parallel^2.
$$
Again, using the chain rule, we can show that:
\begin{equation}
\label{eqn:second-variation}
\delta^2 \text{AELBO}[\lambda,\zeta] = \mathbb{E}\left[\sum_{i=1}^n\sum_{j=1}^n\frac{\partial^2 g(\lambda(Y),Y)}{\partial\lambda_i\partial\lambda_j}\zeta_i(Y)\zeta_j(Y)\right] < -\kappa \parallel \zeta \parallel^2.
\end{equation}

We now show that Eq.~(\ref{eqn:second-variation}) implies that the Hessian of the ELBO (with respect to $\lambda$) is negative definite.
Let $x$ be a vector in $\R^n$ different than zero and $y$ be in the support of the data density.
It suffices to show that:
\begin{equation}
\label{eqn:negative_definite}
\sum_{i=1}^n\sum_{j=1}^n\frac{\partial^2 g(\lambda(y),y)}{\partial\lambda_i\partial\lambda_j}x_ix_j < 0.
\end{equation}
To this end, pick a $\zeta$ whose components are:
$$
\color{black} \zeta_i(Y) = \zeta_{\epsilon,i}(Y) = \frac{x_i {\delta_\epsilon}^{\frac{1}{2}}(Y-y)}{\sqrt{p(Y)}},
$$
\textcolor{black}{where,  $\color{black} \delta_\epsilon(Y-y) = \frac{1}{{|\epsilon| \sqrt{\pi}}} e^{-\frac{{(Y-y)^2}}{{\epsilon^2}}}$, such that $\displaystyle{\lim_{{\epsilon \to 0^{+}}}} \delta_\epsilon(Y-y)= \delta(Y-y).$}
\textcolor{black}{Plugging this $\zeta$ on the left-hand-side of Eq.~(\ref{eqn:second-variation}) we get:}
\begin{equation*}
\begin{split}
\color{black}\mathbb{E}\left[\sum_{i=1}^n\sum_{j=1}^n\frac{\partial^2 g(\lambda(Y),Y)}{\partial\lambda_i\partial\lambda_j}\zeta_i(Y)\zeta_j(Y)\right] &\color{black} = \color{black} \mathbb{E}\left[\sum_{i=1}^n\sum_{j=1}^n\frac{\partial^2 g(\lambda(Y),Y)}{\partial\lambda_i\partial\lambda_j}\zeta_{\epsilon,i}(Y)\zeta_{\epsilon,j}(Y)\right], \\[1pt]
&\color{black}= 
\color{black} \mathbb{E}\left[\sum_{i=1}^n\sum_{j=1}^n\frac{\partial^2 g(\lambda(Y),Y)}{\partial\lambda_i\partial\lambda_j}\frac{x_ix_j\delta_\epsilon(Y-y)}{p(Y)}\right].
\end{split}
\end{equation*}
\textcolor{black}{Similarly, plugging $\zeta$ on the right-hand-side of Eq.~(\ref{eqn:second-variation}) gives:}
$$\color{black}
-\kappa \parallel \zeta\parallel^2 = 
-\kappa \ \mathbb{E}\left[\sum_{i=1}^n\zeta_i^2(Y)\right] = -\kappa \ \mathbb{E}\left[\sum_{i=1}^n\left(\frac{x_i {\delta_\epsilon}^{\frac{1}{2}}(Y-y)}{\sqrt{p(Y)}}\right)^2\right] = -\kappa \ \mathbb{E}\left[\sum_{i=1}^n \frac{x_i^2 \delta_\epsilon(Y-y)}{p(Y)}\right].
$$
\textcolor{black}{From Eq.~(\ref{eqn:second-variation}),}
$$\color{black} \mathbb{E}\left[\sum_{i=1}^n\sum_{j=1}^n\frac{\partial^2 g(\lambda(Y),Y)}{\partial\lambda_i\partial\lambda_j}\zeta_i(Y)\zeta_j(Y)\right] < -\kappa \parallel \zeta \parallel^2,$$
$$\color{black} \implies \color{black} \mathbb{E}\left[\sum_{i=1}^n\sum_{j=1}^n\frac{\partial^2 g(\lambda(Y),Y)}{\partial\lambda_i\partial\lambda_j}\frac{x_ix_j\delta_\epsilon(Y-y)}{p(Y)}\right] < -\kappa \ \mathbb{E}\left[\sum_{i=1}^n \frac{x_i^2\delta_\epsilon(Y-y)}{p(Y)}\right],
$$
$$\color{black} \implies \color{black} \sum_{i=1}^n\sum_{j=1}^n\mathbb{E}\left[\frac{\partial^2 g(\lambda(Y),Y)}{\partial\lambda_i\partial\lambda_j}\frac{x_ix_j\delta_\epsilon(Y-y)}{p(Y)}\right] < -\kappa \ \sum_{i=1}^n \mathbb{E}\left[\frac{x_i^2\delta_\epsilon(Y-y)}{p(Y)}\right].
$$
\textcolor{black}{Taking the limit of both sides as $\displaystyle{{\epsilon \to 0^{+}}}$ we get, }
$$\color{black} \implies \color{black} \sum_{i=1}^n\sum_{j=1}^n\frac{\partial^2 g(\lambda(Y),Y)}{\partial\lambda_i\partial\lambda_j}x_ix_j \le -\kappa \ \sum_{i=1}^n x_i^2,
$$
\textcolor{black}{here, $\sum_{i=1}^n x_i^2$ is a positive number and $x$ is not zero,}
$$\color{black} \implies \color{black} \sum_{i=1}^n\sum_{j=1}^n\frac{\partial^2 g(\lambda(Y),Y)}{\partial\lambda_i\partial\lambda_j}x_ix_j \le -\kappa \ \sum_{i=1}^n x_i^2 < 0,
$$
$$\color{black} \implies \color{black} \sum_{i=1}^n\sum_{j=1}^n\frac{\partial^2 g(\lambda(Y),Y)}{\partial\lambda_i\partial\lambda_j}x_ix_j  < 0.
$$
\end{proof}

\subsection{Choice of the guide and parameterization of the amortization function}
\label{subsec:amortized_net}
\noindent


\textcolor{black}{We explore two distinct guide choices for parameterizing the posteriors in this work. 
The first involves representing the posterior as a distribution from a tractable family, utilizing a full-rank multivariate Gaussian. 
The second entails parametrizing the posterior as a distribution from an intractable family, employing a conditional flow guide.}

We use a neural network to represent the amortization function $\lambda(y)$.
More specifically, we write $\lambda = \lambda(y; \phi)$, where $\phi$ are the network parameters to be learned.
We refer to this neural network as the amortization network,
and to $\phi$ as the amortization parameters.

\subsubsection{Gaussian guide:}
\label{subsubsec:Gaussian guide}
\noindent

\textcolor{black}{For parametrizing the posterior with a full-rank multivariate Gaussian guide, we consider} that the amortization network $\lambda(y;\phi)$ has two multi-dimensional outputs, i.e.,
$$
\lambda(y;\phi) = \left(\mu(y;\phi), L(y;\phi)\right).
$$
The first component, $\mu(y;\phi)$ is a $d$-dimensional vector and the second component, $L(y;\phi)$, is a $d\times d$ matrix.
To be specific, $\mu(y;\phi)$ is the mean vector, and $L(y;\phi)$ is the Cholesky factor of the covariance matrix of the multivariate Gaussian on $\xi$:
$$
q_{\lambda(y;\phi)}(\xi)= \mathcal{N}\left(\xi\middle|\mu(y;\phi), \Sigma(y;\phi) \right),
$$
where $\Sigma(y;\phi)=L(y;\phi) L(y;\phi)^T$.

There are no constraints on the $\mu(y;\phi)$ output of the network.
But the Cholesky factor output $L(y;\phi)$ must be lower triangular with positive diagonal~\cite{lutter2019deep}.
We honor these constraints by composing $\lambda(y;\phi)$ from three distinct neural networks $\lambda_1(y;\phi_1),\lambda_2(y;\phi_2)$ and $\lambda_3(y;\phi_3)$.
The first two networks have output dimension $d$ and the third network has output dimension $\frac{d^2-d}{2}$.
All these networks have similar structure (feed-forward networks with ReLU activations), the complete details of which we provide in the numerical examples section (Sec.~\ref{sec:results}).
The first and the third networks end with a linear activation and correspond, respectively, to the mean vector $\mu(y;\phi)$ and the lower triangular part of $L(y;\phi)$.
The second network corresponds to the diagonal of $L(y;\phi)$ and ends with a softplus activation to ensure the positivity constraint.

\textcolor{black}{
However, in cases where the true posterior is non-Gaussian and multimodal, this Gaussian guide will not serve as an effective approximation. 
In such scenarios, flow-based models act as an efficient alternative to parameterize these arbitrary posteriors.
Specifically, standard normalizing flow guides can be utilized to parameterize arbitrary posteriors of a given dataset.
These guides can be further extended to simultaneously learn the posteriors of all datasets using an advanced version called conditional normalizing flow guides.
We delve into these in the following sections.}

\subsubsection{Normalizing flows:}
\label{subsubsec:Normalizing flows}
\noindent

\begin{figure}[H]
    \centering
    \includegraphics[height=1.7in, width=6.2in]{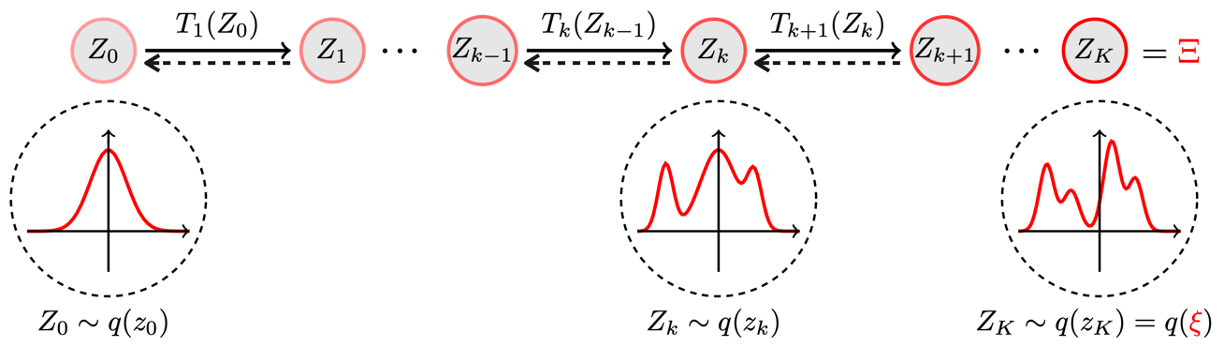}
    \caption{Illustration of a normalizing flow model, transforming a simple distribution to a complex one step by step.}
    \label{fig:SNF}
\end{figure}

\textcolor{black}{Normalizing flows are a class of generative models, wherein a sequence of invertible transformation functions transform a simple distribution into an arbitrary complex distribution as illustrated in the Fig.~\ref{fig:SNF}.}

\textcolor{black}{
The complex distribution $q(\xi)$ in the end can be obtained from a simple distribution $q(z_0)$ through the following procedure.
For simplicity, let us begin by establishing the relationship between the random variables $Z_{k-1}$ and $Z_k$. 
To initiate the process, we sample $z_{k-1}$ from $q(z_{k-1})$.
Subsequently, we consider a bijective transformation $T_k(z_{k-1})$ that maps the random variable $Z_{k-1} \in \mathcal{R}^d$  to $Z_{k} \in \mathcal{R}^d$ and then we can compute the distribution $q(z_{k})$ using the change of variables formula as follows:}
$$
q(z_{k}) = q(z_{k-1}) {\bigg | \det\bigg( \frac{\partial z_{k}}{\partial z_{k-1}} 	\bigg) \bigg | }^{-1}, \ \text{where} \ z_{k} = T_k(z_{k-1})
$$

\textcolor{black}{Note here, the transformation $T_k$ has to be invertible so that every sample from $q(z_{k-1})$ is uniquely mapped to the sample in $q(z_{k})$.
Also, the determinant of Jacobian should be easy to evaluate, and the dimensions of $Z_{k-1}$ and $Z_k$ need to be the same. 
Extending the above equation, the arbitrarily complex distribution $q(\xi)$ can be evaluated by pushing $z_0$, a sample from a simple distribution $q(z_0) = N(0, I)$, through a composition of several of these invertible transformation functions as shown in Fig.~\ref{fig:SNF}, so that the distribution obtained at the end is arbitrarily complex.
Hence, we have:}
\begin{equation}
\label{eqn: T_NFs}
\begin{split}
        z_0 \sim q({z}_0)  = N (0, I),\\
        T({z}_0) = T_K \circ T_{K-1} \ldots T_k \circ T_{k-1}  \ldots \circ T_{1} ({z}_0).
\end{split}
\end{equation}

\begin{figure}[H]
    \centering
    \includegraphics[height=0.4in, width=2.25in]{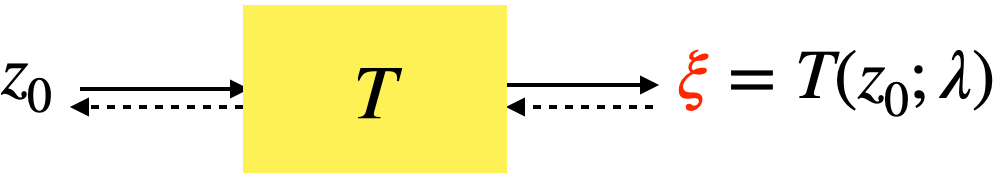}
    \caption{Normalizing flow model's bijective transformation function.}
    \label{fig:SNF_model}
\end{figure}

\textcolor{black}{In this context, the individual bijective transformation functions $T_k$'s are referred to as flows and the composition $T$ is a normalizing flow.
$T_k$'s are parametrized using invertible neural net layers.
Various kinds of invertible neural net architectures are available, establishing bijections with easily computable Jacobian determinants; refer to \cite{kobyzev2020normalizing} for a detailed review.
The parameters encompassing all the flow layers, $\lambda$, are deliberately omitted in Eq.~(\ref{eqn: T_NFs}), see Fig.~\ref{fig:SNF_model}.
Note, that the invertibility of $T({z}_0)$ is guaranteed since the composition of invertible transformation functions is itself invertible. 
Consequently, based on the change of variables, the resulting distribution in the end is given by:}
$$
q_{\lambda}(\xi) = q(z_{0}) {\bigg | \det\bigg( \frac{\partial \xi}{\partial z_{0}} 	\bigg) \bigg | }^{-1}, \ \text{where} \ \xi = T(z_{0};\lambda).
$$ 


\textcolor{black}{Expanding on this, we can
learn the posteriors of all datasets in one shot by following our idea of parametrizing flow parameters, $\lambda$, as a function of the data i.e., $\lambda(y;\phi)$, resulting in $\xi = T(z_{0};\lambda(y;\phi))$. 
In this context, $\lambda(y;\phi)$ is a separate neural network with parameters as $\phi$, which takes in data as input and returns flow parameters as output, these kind of networks are commonly referred to as hypernetworks, refer to \cite{ha2016hypernetworks, chauhan2023brief}.
While this approach works, a more convenient way to learn the posteriors of all datasets is by directly parameterizing $\xi = G(z_{0};y, \phi)$, where $G$ represents an invertible transformation from $Z_0$ to $\Xi$ with respect to $Y$, and $\phi$ are the parameters. 
Architectures adopting this paradigm are known as Conditional normalizing flow guides (or Conditional invertible networks, see \cite{ardizzone2019guided}), which we describe in the subsequent section.}

\subsubsection{Conditional normalizing flow guide:}
\label{subsubsec:Conditional normalizing flow guide}
\noindent

\begin{figure}[H]
    \centering
    \includegraphics[height=1.7in, width=6.2in]{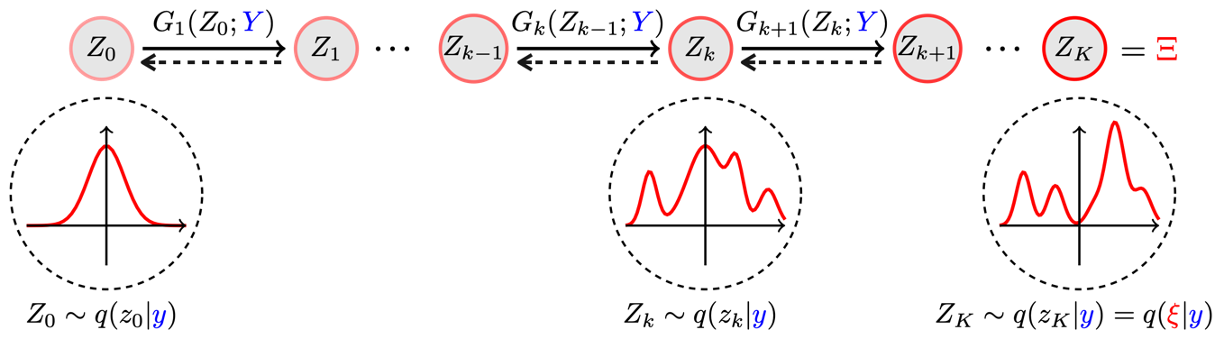}
    \caption{Illustration of a conditional normalizing flow model, transforming a simple conditional distribution to a complex conditional distribution step by step.}
    \label{fig:CNF}
\end{figure}

\textcolor{black}{Conditional normalizing flow models are simple extensions of standard normalizing flows discussed in the previous section (Sec.~\ref{subsubsec:Normalizing flows}). 
Here a sequence of invertible transformation functions, conditioned on the data, transform a simple conditional distribution, $q(z_0|y)$, into an arbitrarily complex conditional distribution, $q(\xi|y)$, as depicted in the Fig.~\ref{fig:CNF}.}

\textcolor{black}{To illustrate, under conditional flow models, a random variable $Z_{k-1} \in \mathcal{R}^d$ is transformed to $Z_k \in \mathcal{R}^d$ by a bijective transformation function, $G_k(Z_{k-1};Y)$, that is dependent on the data. 
The resulting distribution of the transformed variable is given by:}
$$
q(z_{k}|y) = q(z_{k-1}|y) {\bigg | \det\bigg( \frac{\partial z_{k}}{\partial z_{k-1}} 	\bigg) \bigg | }^{-1}, \ \text{where} \ z_{k} = G_k(z_{k-1};y),
$$
\textcolor{black}{where, the transformation between $Z_{k-1}$ and $Z_{k}$ has to be invertible with respect to $Y$, so that every sample from $q(z_{k-1}|y)$ is uniquely mapped to the sample in $q(z_{k}|y)$.
Similar to standard flows, the determinant of the Jacobian should be computationally inexpensive to evaluate. 
Building on the above equation, the resulting arbitrary complex distribution, $q(\xi|y)$, is obtained by pushing a sample from a simple base conditional distribution through a composition of several invertible transformation functions.
The complex distribution in the end is obtained as follows:}
$$
z_0 \sim q({z}_0|y)  = N (0, I),
$$
$$
G({z}_0;{y}) = G_K \circ G_{K-1} \ldots G_k \circ G_{k-1}  \ldots \circ G_{1} ({z}_0;{y}),
$$
$$
q(\xi|y) = q(z_{0}|y) {\bigg | \det\bigg( \frac{\partial \xi}{\partial z_{0}} 	\bigg) \bigg | }^{-1}, \ \text{where} \ \xi = G(z_{0};y).
$$ 

\begin{figure}[H]
\centering
\subcaptionbox{Forward pass.\protect\label{fig:CCB_forwardpass}}%
{\includegraphics[height=1.7in, width=4.7in]{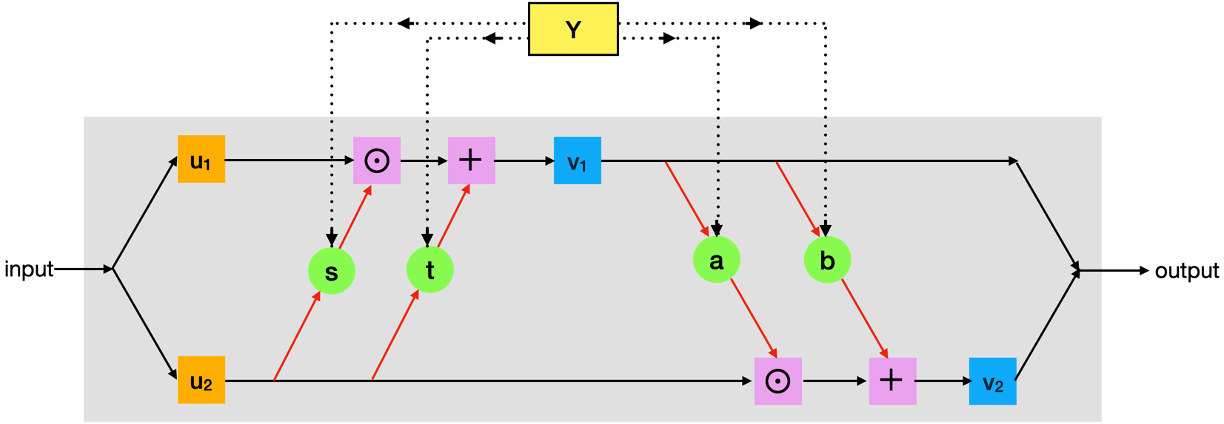}}\\
\subcaptionbox{Inverse pass.\protect\label{fig:CCB_inversepass}}%
{\includegraphics[height=1.7in, width=4.7in]{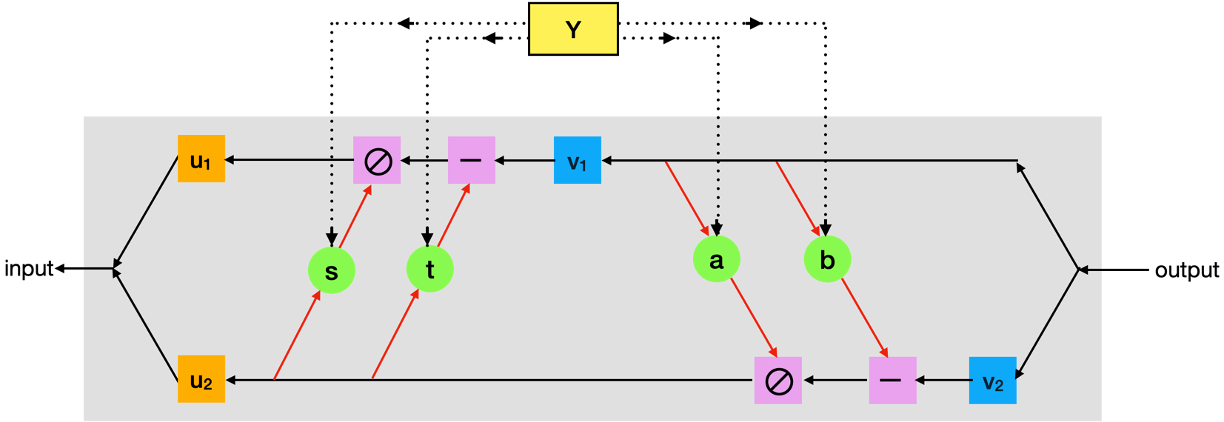}}
\caption{One conditional coupling block.}
\label{fig:CCB}
\end{figure}

\textcolor{black}{
In this work, we choose the bijective transformation function $G_k$ to be a conditional coupling block (CCB) \cite{ardizzone2019guided} followed by a fixed permutation operation, i.e., $ G_k = \widehat{G_k} \circ P_k$. 
Here, $ \widehat{G_k} : \mathbb{R}^d \to \mathbb{R}^d$ represents an invertible nonlinear transformation defined by CCB (see Fig.~\ref{fig:CCB}). 
The CCB involves four distinct neural networks denoted as `$s$', `$t$', `$a$', and, `$b$'. 
Let the input vector of $\widehat{G_k}$ be $u$ and the output vector be $v$, such that $v = \widehat{G_k} (u; y)$.
Invertibility of the CCB is achieved by splitting the input vector into two halves, $u_1$ and $u_2$, and then by performing the following operations on the input:}
\begin{equation*}
\begin{array}{ccc}
z_{k-1} = u,\\
u = [u_1, u_2],\\
u_1 = u_{1:\lfloor{\frac{d}{2}}\rfloor};\  u_2 = u_{{\lfloor{\frac{d}{2}}\rfloor}+1:d},\\
{v}_1 = {u}_1 \odot \exp(s({u}_2, y)) + t({u}_2, y),\\
{v}_2 = {u}_2 \odot \exp(a({v}_1, y)) + b({v}_1, y),\\ 
v = [v_1, v_2],\\
\widehat{G_k} (z_{k-1};y) = v,
\end{array}
\end{equation*}
\textcolor{black}{where, $\odot$ is the Hadamard product or element-wise product (see Fig.~\ref{fig:CCB_forwardpass}).
This formulation ensures that the Jacobian of the transformation is triangular, and that the determinant is computationally efficient to evaluate. 
The determinant is given by  ${\displaystyle \prod_{j=1}^{\lfloor{\frac{d}{2}}\rfloor} [\exp(s({u}_2, y))]_j}  {\displaystyle \prod_{j=1}^{d-\lfloor{\frac{d}{2}}\rfloor} [\exp(a({v}_1, y))]_j}$.
The outputs $v_1$ and $v_2$ are concatenated as $v = [v_1, v_2]$, and then passed through a fixed permutation layer.
The permutation layer shuffles the elements of $v$ in a randomized but fixed manner before passing into further stages of computation.
Note that, these permutation operations are volume-preserving and the modulus of their determinant is $1$ i.e.,  $|\det P_k| = 1 \ \forall \ k=1 \text{ to } K.$ 
However, at the end of the flow, the permutation operation is not applied as there is no further computation i.e., $P_K = I$.
These permutation operations help to make the transformation expressive enough to represent a complex nonlinear transformation.
Note also that the inverse mapping is trivially available by using a CCB because of the affine operations in it. 
The inverse mapping is given by (see Fig.~\ref{fig:CCB_inversepass}):}
\begin{equation*}
\begin{array}{ccc}
\widehat{G_k} (z_{k-1};y) = [v_1, v_2],\\
{u}_2 = ({v}_2 - b({v}_1, y)) \oslash \exp(a({v}_1, y)),\\ 
{u}_1 = ({v}_1 - t({u}_2, y)) \oslash \exp(s({u}_2, y)),\\ 
z_{k-1} = [u_1, u_2],
\end{array}
\end{equation*}
\textcolor{black}{where, $\oslash$ is the element-wise division.}

\textcolor{black}{Moreover, the inner functions `$s$', `$t$', `$a$', and, `$b$' can be arbitrary neural networks i.e., a fully connected network, or a CNN, and need not be invertible as they are only ever estimated in the forward direction during both forward and inverse pass through CCB, i.e., the arrows in red are pointed in the same direction both in Fig.~\ref{fig:CCB_forwardpass} and Fig.~\ref{fig:CCB_inversepass}. 
We choose these inner functions to be feed-forward neural networks. 
Moreover, we pass in the observed data, $y$, as input to these functions directly. 
However, we can also pass these observed data through some other neural network first and then we can pass the transformed observed data to these inner functions.}

\begin{figure}[H]
    \centering
    \includegraphics[height=1.25in, width=2.8in]{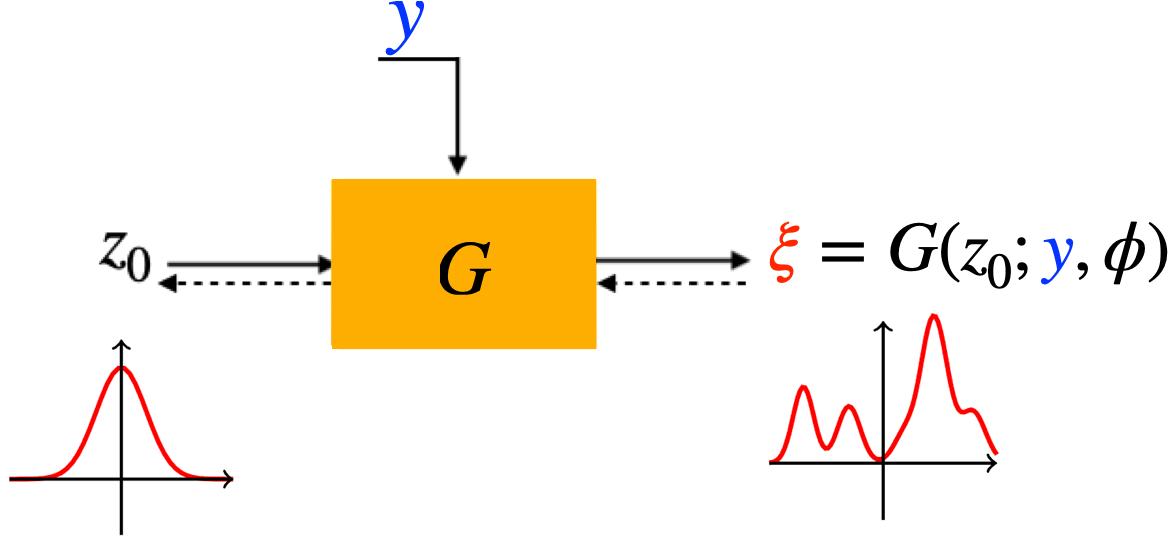}
    \caption{Conditional normalizing flow model's transformation function.}
    \label{fig:CNF_model}
\end{figure}

\begin{figure}[H]
    \centering
    \includegraphics[height=1.25in, width=5.5in]{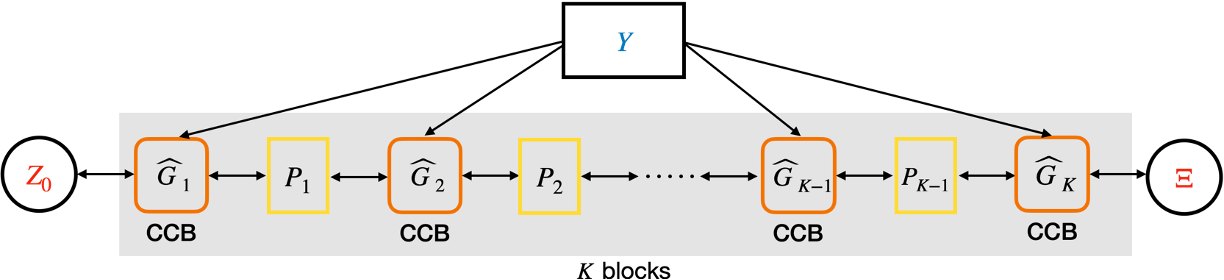}
    \caption{Detailed schematic of conditional normalizing flow model of length $K$.}
    \label{fig:Amortization_net_CNF}
\end{figure}

\textcolor{black}{In this way, we compose several CCBs and fixed permutation layers one after the other to represent a complex approximate posterior.
This composition can be looked at as a function parameterized with trainable network parameters as $\phi$ i.e., $\Xi = G(Z_0; Y,\phi).$
In this context, we refer to this network as the amortization
network and $G$ as conditional normalizing flow (see Fig.~\ref{fig:CNF_model} and Fig.~\ref{fig:Amortization_net_CNF}).}

\subsection{Numerical optimization via stochastic gradient ascent}
\label{subsec:optimization}
\noindent

\textcolor{black}{The above parameterizations in Sec.~\ref{subsubsec:Gaussian guide}, and, Sec.~\ref{subsubsec:Conditional normalizing flow guide} defines a subset of the admissible amortization functions $\mathcal{A}$.}
This subset is described by $\phi$ which lives in an unrestricted Euclidean space.
From now on, we seek to solve the finite-dimensional optimization problem of maximizing the multivariate function:
\begin{equation}
\label{eqn:objective-function}
v(\phi) = \text{AELBO}[\lambda(\cdot;\phi)].
\end{equation}

We construct a stochastic gradient ascent algorithm that provably converges to a local maximum of Eq.~(\ref{eqn:objective-function}).
The first step is to recast the problem as a stochastic optimization problem and to construct an unbiased estimator of the gradients of the objective function $v$ with respect to the amortization network parameters $\phi$.

Notice that the objective function decomposes in two parts.
An expectation over the logarithm of the joint probability density of the data $Y$ and the parameters $\Xi$ and an expectation over the data density of the entropy of the guide:
\begin{equation}
\label{eqn:J_as_exp}
v(\phi) = \mathbb{E}\left[\mathbb{E}_{\Xi\sim q_{\lambda(Y;\phi)}}\left[\log p(\Xi,Y)\middle|Y\right] + \mathbb{H}\left[q_{\lambda(Y;\phi)}(\Xi)\middle|Y\right]\right].
\end{equation}
In this equation, $\mathbb{E}[\cdot|Y]$ is and $\mathbb{H}[\cdot|Y]$ are the expectation and entropy conditional on $Y$, respectively.

\textcolor{black}{We recast this equation into a stochastic optimization problem.
We explore this process in two contexts: first, when adopting a Gaussian guide, and next, when employing a conditional normalizing flow guide.}

\textcolor{black}{When using the Gaussian guide,}
we employ the reparameterization trick~\cite{kucukelbir2017automatic, kingma2013auto, reparameterizationtrick} for the first summand of Eq.~(\ref{eqn:J_as_exp}) to remove the dependence of the expectation on the amortization network parameters.
Introduce the $d$-dimensional standard normal random variable $Z_0\sim N(0, I)$, and write:
$$\Xi = \mu(Y;\phi) + L(Y;\phi)Z_0.$$ 
Then $\Xi$ conditioned on $Y$ follows $q_{\lambda(Y;\phi)}$ and thus:
$$
 \mathbb{E}\left[\mathbb{E}_{\Xi\sim q_{\lambda(Y;\phi)}}\left[\log p(\Xi,Y)\middle|Y\right]\right]
 =  \mathbb{E}\left[\mathbb{E}\left[\log p(\Xi=\mu(Y;\phi) + L(Y;\phi)Z_0,Y)\middle|Y\right]\right]
     = \mathbb{E}[\log p(\Xi=\mu(Y;\phi) + L(Y;\phi)Z_0, Y)].
$$
For the second term in Eq.~(\ref{eqn:J_as_exp}), \textcolor{black}{ we evaluate the entropy of the Gaussian guide, which is given by:}
$$
\mathbb{H}\left[q_{\lambda(Y;\phi)}(\Xi)\middle|Y\right]
= \frac{1}{2} \log \{(2\pi e)^d \det \left(\Sigma(Y;\phi)\right)\}
= \frac{d}{2}\log (2\pi e) + \sum_{r=1}^d \log L_{rr}(Y;\phi).
$$
Putting everything together, we get:
$$
v(\phi) = \frac{d}{2}\log (2\pi e)+\mathbb{E}\left[\log p(\Xi=\mu(Y;\phi) + L(Y;\phi)Z_0, Y)+\sum_{r=1}^d \log L_{rr}(Y;\phi)\right].
$$

The reparameterization trick allows us to derive unbiased estimators of $v(\phi)$ and of its gradient with respect to $\phi$.
To this end, let $N_y$ and $N_z$ be integers.
Let $Y_i$, $i=1,\dots, N_y$, be independent identically distributed (iid) random variables following the data density.
Let $Z_{0j}$, $j=1,\dots,N_z$, be iid following a $d$-dimensional standard normal.
Define the random variable:
\begin{equation}
\label{eqn:V_function_Gaussianguide}
V(\phi) = \frac{d}{2}\log (2\pi e)+\frac{1}{N_y}\sum_{i=1}^{N_y}\left\{\frac{1}{N_z}\sum_{j=1}^{N_z}\log p(\Xi=\mu(Y_i;\phi) + L(Y_i;\phi)Z_{0j}, Y_i)+\sum_{r=1}^d \log L_{rr}(Y_i;\phi)\right\}.
\end{equation}

\textcolor{black}{When we employ a conditional flow guide, we define the random variable $V(\phi)$ as outlined below.
Here we know that:}
$$\Xi = G(Z_0; Y, \phi),$$
\textcolor{black}{so, the first and second terms of Eq.~(\ref{eqn:J_as_exp}) reduces to:}
$$
 \mathbb{E}\left[\mathbb{E}_{\Xi\sim q_{\lambda(Y;\phi)}}\left[\log p(\Xi,Y)\middle|Y\right]\right]
     = \mathbb{E}[\log p(\Xi=G(Z_0;Y,\phi), Y)],
$$
$$
 \mathbb{E}\left[\mathbb{H}\left[q_{\lambda(Y;\phi)}(\Xi)\middle|Y\right]\right]
     = \frac{d}{2}\log (2\pi e) + \mathbb{E}\left[ \log \bigg | \det\bigg( \frac{d(\Xi=G(Z_0;Y,\phi))}{dZ_{0}} 	\bigg) \bigg | \right],
$$
\textcolor{black}{say, $ \bigg | \det\bigg( \frac{d(\Xi=G(Z_0;Y,\phi))}{dZ_{0}} \bigg) \bigg | = h(Z_0;Y,\phi)$, then we have $v(\phi)$ as:}
$$
v(\phi) = \frac{d}{2}\log (2\pi e) + \mathbb{E}\left[\log p(\Xi=G(Z_0;Y,\phi), Y) + \log(h(Z_0;Y,\phi)) \right] .
$$
\textcolor{black}{Following this, the random variable $V(\phi)$ is:}
\begin{equation}
\label{eqn:V_function_CNFguide}
V(\phi) = \frac{d}{2}\log (2\pi e)+\frac{1}{N_y}\frac{1}{N_z}\sum_{i=1}^{N_y}\sum_{j=1}^{N_z}\left\{ \log p(\Xi=G(Z_{0j};Y_i,\phi), Y_i) + \log(h(Z_{0j};Y_i,\phi)) \right\},
\end{equation}
\textcolor{black}{here, $Z_{0j}$, $j=1,\dots, N_z$ and $Y_i$, $i=1,\dots, N_y$ are iid $d$-dimensional standard normal random variables and random variables following the data density, respectively.}

\textcolor{black}{When employing either Gaussian or conditional flow guides, for the random variable $V(\phi)$ in Eq.~(\ref{eqn:V_function_Gaussianguide}) or Eq.~(\ref{eqn:V_function_CNFguide}), we have:}
$$
v(\phi) = \mathbb{E}\left[V(\phi)\right].
$$
We have now succeeded in recasting the problem of learning the amortization network parameters as a stochastic optimization problem:
$$
\phi^* = \arg\max_{\phi} v(\phi) = \arg\max_{\phi}\mathbb{E}[V(\phi)].
$$
Furthermore, the 
$$
\nabla_{\phi} v(\phi) = \mathbb{E}\left[\nabla_{\phi}V(\phi)\right],
$$
where $\nabla_{\phi}$ denotes the gradient with respect to $\phi$.

Under these conditions, the stochastic gradient ascent updates:
\begin{equation}
\label{eqn:adams_update}
\phi_{k+1} = \phi_k + \eta_k \nabla_{\phi} v_k(\phi_k),
\end{equation}
where $v_k(\phi_k)$ are independent samples from $V(\phi_k)$ (which can be constructed by sampling the underlying $Y_i$'s and $Z_{0j}$'s)
converge to a local maximum if the learning rate $\eta_k$ satisfies the Robbins-Monro conditions~\cite{robbins1951stochastic}:
$$
\sum_{k=1}^\infty \eta_k = +\infty,
$$
and
$$
\sum_{k=1}^\infty \eta_k^2 < +\infty.
$$
This algorithm is typically called stochastic gradient ascent (SGA).

In our numerical examples, we employed the adaptive moments (ADAM) optimization method~\cite{kingma2014adam}, a robust variant of SGA that typically exhibits faster convergence.
This method computes adaptive learning rates for each parameter using exponentially decaying averages of past gradients and past squared gradients and converges faster than SGA. 
In ADAM, the averaging hyper-parameters denoted as $\beta_1$, and $\beta_2$  are, in principle, tunable.
In practice, default values of $\beta_1=0.9$, $\beta_2=0.999$, as suggested by~\cite{kingma2014adam} work well and we do not change these quantities. 
We use a step decay learning rate schedule, which decays the learning rate by a multiplicative factor $\alpha$ after every $r$  iterations.

\begin{figure}[H]
    \centering
    \includegraphics[height=3.4in, width=6.0in]{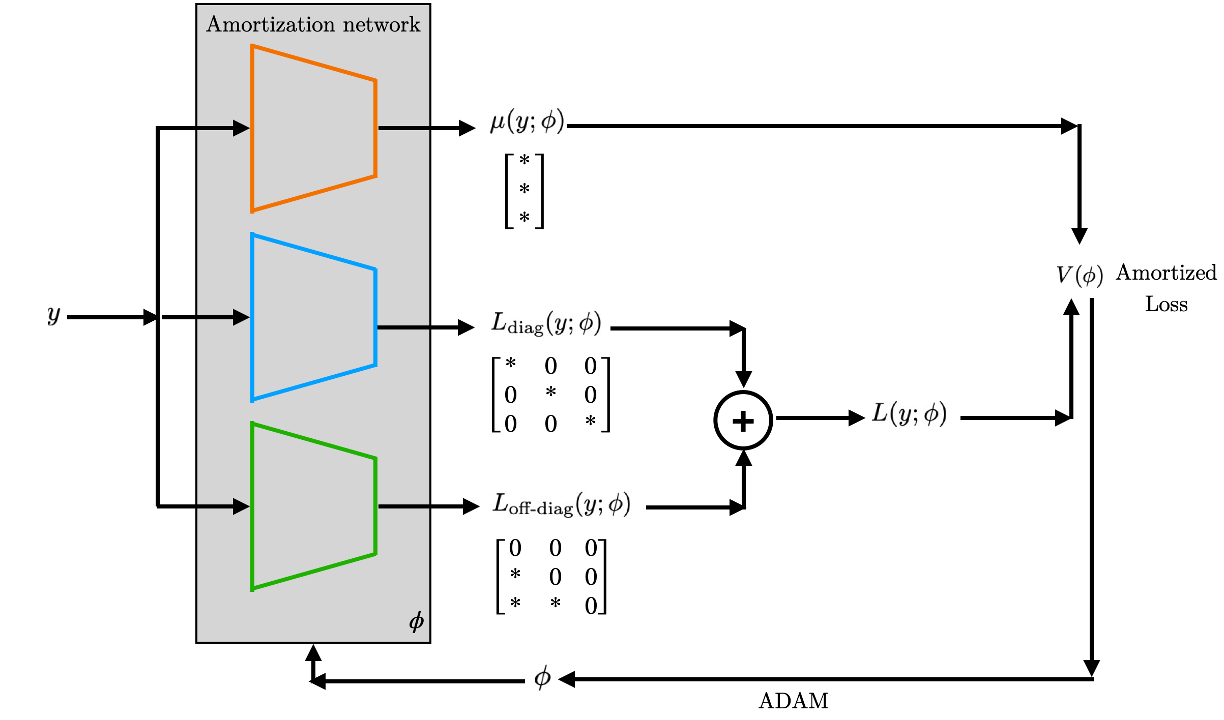}
    \caption{
    Schematic representation of the amortization approach for learning Bayesian inverse maps \textcolor{black}{with full-rank Gaussian guide}. 
    The amortization network in grey takes in observation data and outputs the corresponding variational parameters of full-rank Gaussian distribution. Shown in orange, blue, and green are the three neural networks computing $\mu(y;\phi), L_{\text{diag}}(y;\phi)$ and $ L_{\text{off-diag}}(y;\phi)$ respectively.  
    }
    \label{fig:Schematic_of_Amortization_network_Gaussian_guide}
\end{figure}

\begin{figure}[H]
    \centering
    \includegraphics[height=1.8in, width=4.5in]{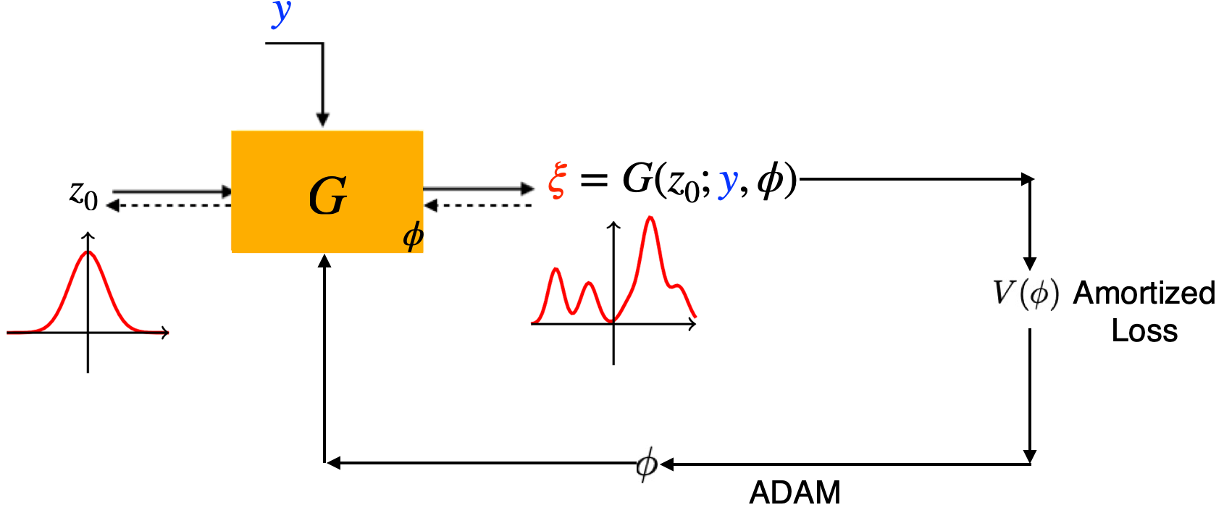}
    \caption{
    \textcolor{black}{Schematic representation of the amortization approach for learning Bayesian inverse maps with Conditional normalizing flow guide. 
    The amortization network in orange takes in observation data and samples from a simple base conditional as input and outputs the corresponding posterior samples.}
    }
    \label{fig:Schematic_of_Amortization_network_CNF_guide}
\end{figure}

The amortization network training algorithm is outlined in Algorithm~\ref{alg:Inference network training}. 
\textcolor{black}{Fig.~\ref{fig:Schematic_of_Amortization_network_Gaussian_guide}  and Fig.~\ref{fig:Schematic_of_Amortization_network_CNF_guide} shows a schematic representation of the proposed approach with a full-rank multivariate Gaussian guide and conditional normalizing flow guide, respectively.}

\begin{algorithm}
\caption{Amortization network training process} \label{alg:Inference network training}
\begin{algorithmic}[1]
\Require Amortization network architecture, number of iterations $N_{\text{iter}}$, number of samples $(N_{y}, N_{z})$, initial learning rate $\eta_{0}$, multiplicative factor for learning rate decay $\alpha$, learning rate decay after every $r$ iterations.
\State Initialize parameters of the amortization network. 
\For{$k$ = 1 to $N_{\text{iter}}$}
\For {\text{each} $i = 1, \cdots, N_{y}$} 
\State Generate sample data $y_{ki}$ by sampling the random variable $Y_i$ that follows the data density by:
\State \indent Sampling parameters $\xi_{ki}$ from the prior $p(\xi)$.
\State \indent Solving the forward model $f(\xi_{ki})$.
\State \indent Sampling $y_{ki}$ from the likelihood.
\For {\text{each} $j = 1, \cdots, N_{z}$} 
\State Generate samples $z_{kj}$ of the standard normal $Z_j$.
\EndFor
\EndFor
\State \textcolor{black}{Construct a sample $v_k(\phi_k)$ of $V(\phi_k)$ using Eq.~(\ref{eqn:V_function_Gaussianguide}) or Eq.~(\ref{eqn:V_function_CNFguide}) depending on the guide chosen.}
\State Construct a sample $\nabla_{\phi}v_k(\phi_k)$ of $\nabla V(\phi_k)$.
\State Update the learning rate $\eta_k$ based on step decay learning rate schedule using $\eta_{0}$, $\alpha$ and $r$.
\State Update the parameters $\phi_{k+1}$ using Eq.~(\ref{eqn:adams_update}).
\EndFor
\\
\Return $\phi^{*}$. \Comment{Return $\phi^{*}$ trained amortization network parameters.}
\end{algorithmic}
\end{algorithm}

\section{Numerical examples}
\label{sec:results}
\noindent

We demonstrate the effectiveness of our proposed framework to learn inverse maps through three examples. 
These examples allow intuitive visualizations of the posterior and comparison against ground truth MCMC estimates of the posterior.
The MCMC estimates are sampled by employing the No-U-Turn sampler
(NUTS)~\cite{hoffman2014no} implemented by the 
Pyro~\cite{bingham2019pyro} python library.
The codes of all the examples in this work are published in Github (\url{https://github.com/PredictiveScienceLab/paper-2023-inverse-map-karumuri}).
\textcolor{black}{For the first two examples, we employ an amortization network with a full-rank multivariate Gaussian guide as in Sec.~\ref{subsubsec:Gaussian guide} to learn the Bayesian inverse map. 
This choice is motivated by the proximity of ground truth MCMC estimates to a Gaussian distribution in these two examples.
Whereas, for the third example, the ground truth MCMC estimates adhere to a complex distribution, hence we opt for a conditional normalizing flow guide as in Sec.~\ref{subsubsec:Conditional normalizing flow guide} to estimate the Bayesian inverse map.
However, for the sake of comprehensive comparison, we also present results obtained with an amortized Gaussian guide for this example.}

\subsection{Comparison metrics}
\noindent

To assess the quality of the inverse map, one of the metrics we use is the value of the Kolmogorov-Smirnov (KS) test statistic~\cite{ref1-kstest, ksmirnov} between the posterior samples estimated by our method and the corresponding samples estimated using the MCMC method. 
This test statistic quantifies the distance between the empirical cumulative distribution functions (ECDFs) of the posterior samples obtained by both methods. The test statistic is zero when the posterior samples of the parameters from both methods follow the same distribution.

We estimate these KS statistic values of posteriors for $N_{y} = 100$ samples from the data density, say $y_i$, $i=1,\dots, N_y$.
For each one of these hypothetical data sets, we perform MCMC to obtain $N_{\text{MCMC}}=1000$ samples of the parameters, say $\xi_{ij}$, $j=1,\dots, N_{\text{MCMC}}$.
Specifically, we use a total of $3{,}300$ NUTS samples, discard the first $300$ as burn-in, and then select every 3\ts{rd} sample.

Another metric we use is the so-called re-simulation error.
Let $\Xi_{\text{gt}}$ be a random variable that follows the prior $p(\xi)$.
The ``gt'' subscript stands for ``ground truth.''
Let $Y$ be the random variable modeling the measurement we would have made if the parameters were $\Xi_{\text{gt}}$, i.e., $Y \sim p(y|\xi=\Xi_{\text{gt}})$.
The re-simulation error is defined to be:
$$
    \mathcal{E}_{\text{re-sim}} = \mathbb{E}\left[\mathbb{E}_{\Xi\sim q_{\lambda(Y;\phi)}}\left[\lVert f(\Xi)-f(\Xi_{\text{gt}})\rVert_2\middle| \Xi_{\text{gt}},Y\right]\right].
$$
Note that the inner expectation is over the guide $q_{\lambda(Y;\phi)}$, while the outer expectation is over the ``ground truth'' parameters $\Xi_{\text{gt}}$ and the hypothetical measurement $Y$.
Again, we approximate the re-simulation error by sampling.
Specifically, we use $N_y$ samples $y_i$ and $\xi_{\text{gt},i}$ of $Y$ and $\Xi_{\text{gt}}$, respectively.
For each $i=1,\dots,N_y=100$, we sample $N_{\text{samples}} = 1000$ points $\xi_{ij}$ from the guide $q_{\lambda(y_i;\phi)}$.
It is:
\begin{equation}
\label{eqn: re-simulation error}
\hat{\mathcal{E}}_{\text{re-sim}} = \frac{1}{N_{y}N_{\text{samples}}} \sum_{i=1}^{N_{y}} \sum_{j=1}^{N_{\text{samples}}}
    \lVert f(\xi_{ij}) - f(\xi_{\text{gt},i}) \rVert_2.
\end{equation}
The benefit of the re-simulation error is that it does not require any MCMC samples.
%
\subsection{Damage location detection}
\noindent

We consider the physically motivated problem of identification of the location and size of the damage in an object using the Electrical impedance tomography (EIT)~\cite{adler2021electrical, brown2003electrical, henderson1978impedance} technique.
This technique is taking center stage over recent years in Structural health monitoring (SHM)~\cite{farrar2007introduction, balageas2010structural} owing to its low cost and non-intrusive nature for damage detection.
In EIT, the test object is probed using low-intensity current injections, and, thereby measuring the induced boundary electric potential.
Changes in the measured electric potential data are linked to the changes in material properties of the test object via an inverse problem approach.

\begin{figure}[H]
     \centering
     \subfloat[Electrical conductivity field of test object.]{\includegraphics[width=0.45\textwidth]{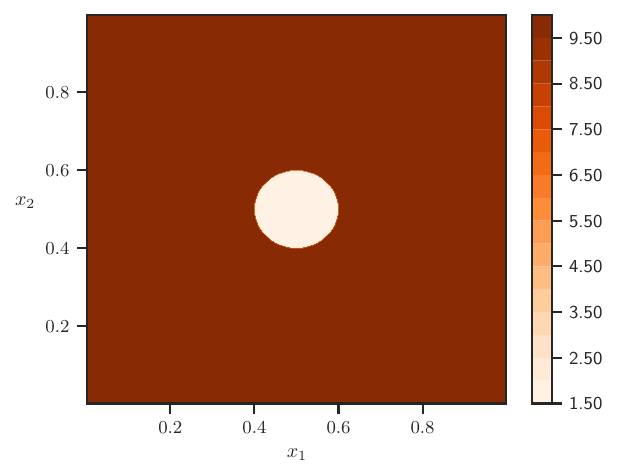}}
     \subfloat[Potential field induced in \textit{Expt-1}.]{\includegraphics[width=0.45\textwidth]{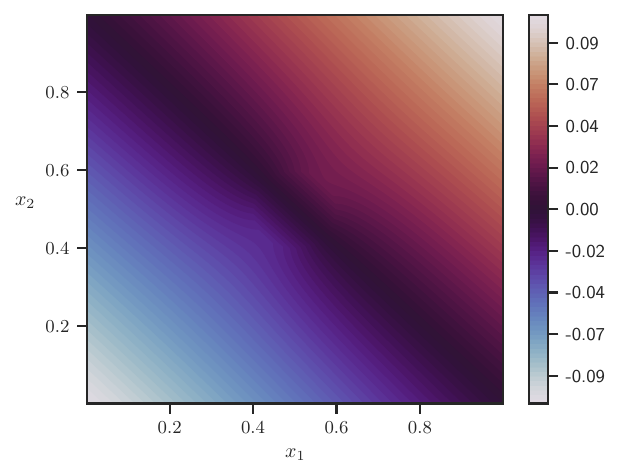}}\\
     \subfloat[Potential field induced in \textit{Expt-2}.]{\includegraphics[width=0.45\textwidth]{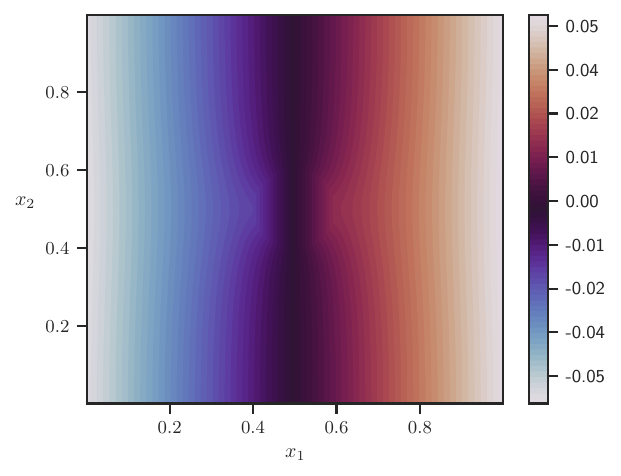}}
     \subfloat[Potential field induced in \textit{Expt-3}.]{\includegraphics[width=0.45\textwidth]{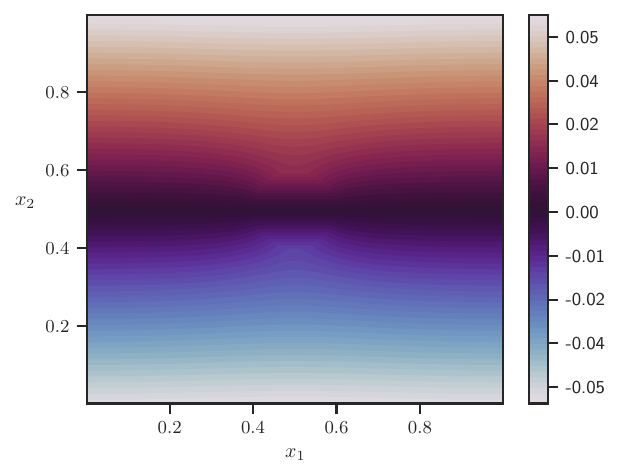}}
     \caption{(Damage location detection) Illustration of the electrical conductivity field and induced potential (FVM solution) in a test object with circular damage at $(0.5, 0.5)$.}
     \label{fig:DD - conductivity and FVM solution}
\end{figure}

In this context of damage identification using the EIT technique, the Bayesian inverse map learned using our AVI approach enables instantaneous on-the-fly detection of the distribution of damage location given the measured boundary potential data.
To demonstrate this, we take a square solid plate of unit length with a circular void of $0.1$ radius and we aim at discovering the void center using the EIT technique.
Mathematically, the forward electrostatic EIT boundary value problem is described by the following equation~\cite{zhao2019detection}:
\begin{equation}
\label{eqn: DD equation}
    - \nabla\cdot\left(a(x)\nabla u(x) \right) = 0,\;\forall\;x\in\Omega=[0, 1]^2\subset\R^{2},
\end{equation}
where $\Omega$ indicates domain of the square plate, $u = $ electric potential, 
$a = $ internal electric conductivity of the material, and the conductivity of the material varies as follows:
\begin{equation}
\begin{gathered}
a(x) =
\begin{cases}
a_{d} &\text{within the circular defect with center at } (x_{1c}, x_{2c})\\
a_{o}&\text{otherwise}\\
\end{cases} 
\end{gathered}
\end{equation}
with $a_{d} = 1.5$ and $a_{o} = 10$. The test object is subjected to Neumann boundary conditions on the boundaries as follows:
\begin{equation}
\begin{gathered}
a_{o} \frac{\partial u}{\partial n}=
\begin{cases}
j(\text{current})&\text{on } S, \ S \subset \partial\Omega\\ 
0 &\text{on } \partial\Omega \text{\textbackslash} \Bar{S}
\end{cases} 
\end{gathered}
\end{equation}
with unit current ($j=1$) injected on the boundary $S$, a subset of boundary of the square plate $\partial\Omega$.
Specifically, we consider three experimental scenarios, in the first experiment unit current is injected on all four sides of the object ($S$),
in the second experiment unit current is injected on the left and right sides of the test object ($S$), and no current on the other two sides and finally in the third experiment, a unit current is injected only on the top and bottom of the test object ($S$).
From these three experiments induced potential on the boundaries is measured.
We call these experiments as \textit{Expt-1}, \textit{Expt-2} and \textit{Expt-3}.
For the sake of illustration, we show the contours of induced potential in the three experiments for a test object with circular defect at center $(0.5, 0.5)$ in  Fig.~\ref{fig:DD - conductivity and FVM solution}.
We could clearly see a change in induced potential at the location of the defect.
These induced potentials are estimated by solving the forward EIT problem in Eq.~(\ref{eqn: DD equation}) numerically using a finite volume method (FVM) solver implemented in  \texttt{FiPy}~\cite{guyer2009fipy}.

We assumed that the circular void of radius $0.1$ lies anywhere within the square plate in the region $[0.2, 0.8]^2$ and that we have access to $600 \ (=m)$ noisy measurements of the induced boundary potential in total from the three experiments considered.
To be specific, $200$ measurements from each of the experiments i.e., $50$ measurements on each side of the square plate.
We collectively denote the noisy boundary potential measurements $\{y_{1,1}, y_{1,2},\dots,y_{1, 200}\}$ from \textit{Expt-1} as vector $y_1$. 
Similarly, we denote the data collected from \textit{Expt-2} and \textit{Expt-3} as $y_2$ and $y_3$ respectively.
Now the inverse problem we are interested in here is to identify the center of the circular damage $\xi=\{x_{1c}, x_{2c}\}$ based on the observed boundary potential data from the three experiments $y = \{y_1, y_2, y_3\}$.
We learn the required Bayesian inverse map from all the boundary potential data $y \in \R^{600}$ to the center of the circular damage $\xi \in \R^{2}$ using our amortization network \textcolor{black}{for Gaussian guide as in Sec.~\ref{subsubsec:Gaussian guide}}.

For learning the amortization network, we set up the required likelihood and prior. 
We assume the $m$ noisy  measurements to be independent and identically distributed (iid) and  
we model the measurement process using a Gaussian likelihood with a noise scale of $\gamma=0.005$ :
$$
{p(y |\xi)} = \prod_{i=1}^{3}N(y_i|u_i(\xi),\gamma^2 I),
$$
where, $u_1, u_2, u_3$ are vectors of true boundary potentials from the three experiments which are obtained using the FVM solver.
Note that, each of these vectors is of length $200$.
Further to make the computations faster, for each of the experiments we built a surrogate of true boundary potentials using a residual neural network.
The network takes in the center of the circular void, $\xi$ in $\R^2$ as input, and outputs the corresponding boundary potentials in $\R^{200}$. 
The architecture of this network consists of $5$ residual blocks each with $3$ layers having $60$ neurons each and with SiLU activation functions.
We trained this network using $3{,}721$ data points by randomly generating the circular defects of radius $0.1$ within the region $[0.2, 0.8]^2$ and estimating the corresponding boundary potential induced using the FVM solver mentioned before.
Having learnt these surrogates, now the likelihood above reduces to,
\begin{equation}
{p(y |\xi)} = \prod_{i=1}^{3}N(y_i|\hat{u}_i( \xi;\theta_i),\gamma^2 I),
\end{equation}
where $\theta_i$ are corresponding residual network parameters.
We choose the parameters prior as $\xi \sim \mathcal{N}(\mu, \sigma^2I)$ with $\mu= [0.5, 0.5]$ and $\sigma= [0.1, 0.1]$.

Having obtained the necessary ingredients likelihood and prior, we built the three networks in our amortization net using feed-forward networks.
We consider these networks to be having two hidden layers of sizes 20, and 10 respectively. 
Following Algorithm~\ref{alg:Inference network training} the amortization net is trained for $8{,}000$ iterations ($N_{iter}$) using  $(N_{y}=32, N_{z} =5)$ samples in each iteration. 
Followed by an initial learning rate of  $\eta_{I_0}=\num{e-2}$, with step decay learning rate schedule of multiplicative factor $\alpha=0.1$, after every $r=4{,}000$ iterations.

\begin{figure}[H]
        \centering
        \includegraphics[width=0.5\textwidth]{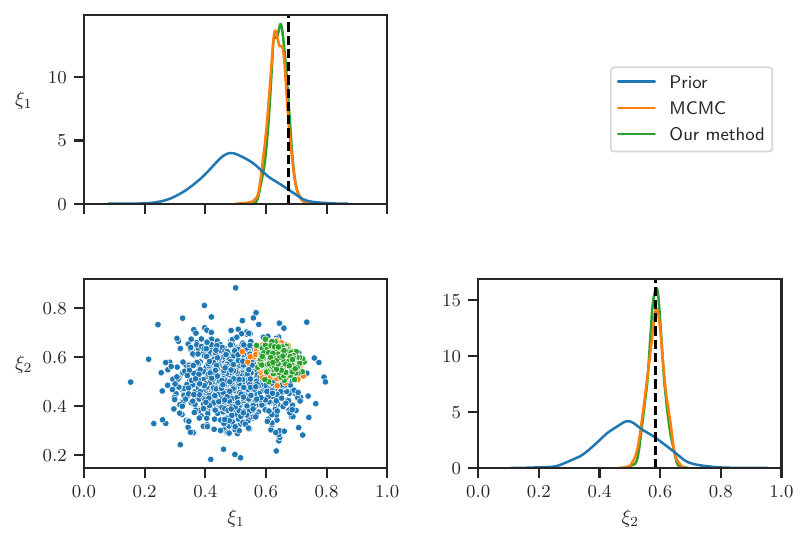}
        \caption{(Damage location detection - Observation set 1.) Qualitative results of the damage detection problem from our method and MCMC approaches using pairplot for the case where ground truth damage center is located at $(0.67, 0.59)$.}
        \label{fig:DD1}
\end{figure}

\begin{figure}[H]
        \centering
        \includegraphics[width=0.5\textwidth]{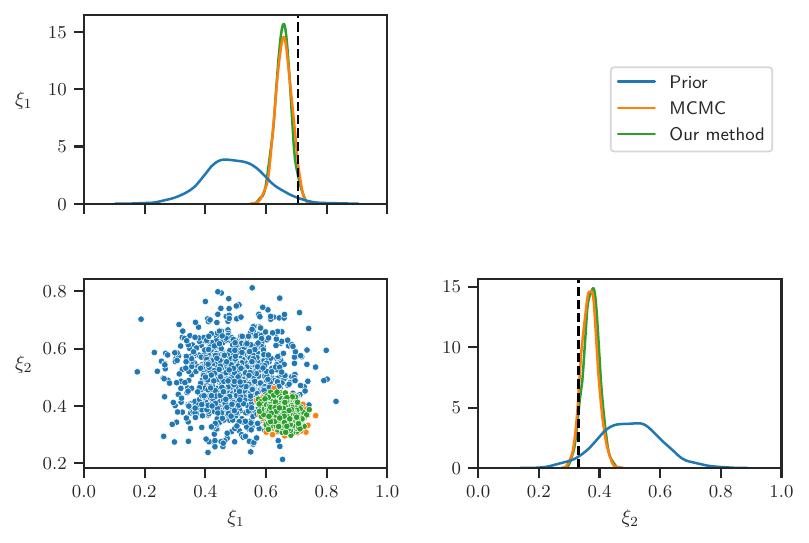}
        \caption{(Damage location detection - Observation set 2.) Qualitative results of the damage detection problem from our method and MCMC approaches using pairplot for the case where ground truth damage center is located at $(0.71, 0.33)$.}
        \label{fig:DD2}
\end{figure}

\begin{figure}[H]
        \centering
        \includegraphics[width=0.5\textwidth]{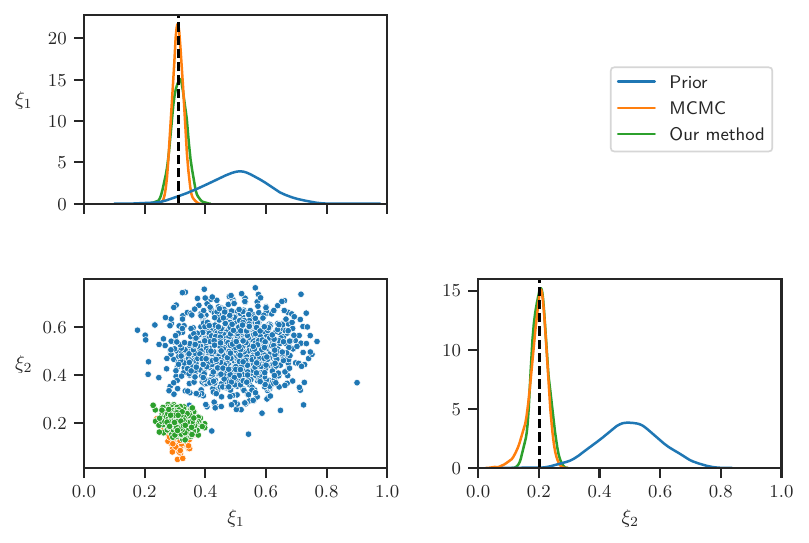}
        \caption{(Damage location detection - Observation set 3.) Qualitative results of the damage detection problem from our method and MCMC approaches using pairplot for the case where ground truth damage center is located at $(0.31, 0.20)$.}
        \label{fig:DD3}
\end{figure}

Qualitative results of the posteriors learned using our amortization net for three sets of observations are shown in Figs.(\ref{fig:DD1} - \ref{fig:DD3}), along with comparisons against corresponding MCMC estimates.
In these figures, diagonal elements show the marginal posterior estimates of the damage location center coordinates and the off-diagonal elements show the corresponding scatter-plot.
The ground truth location of the damage center is shown by a black dashed line for reference on the diagonal elements.

\begin{figure}[t] 
     \centering
     \subfloat{\includegraphics[width=0.45\textwidth]{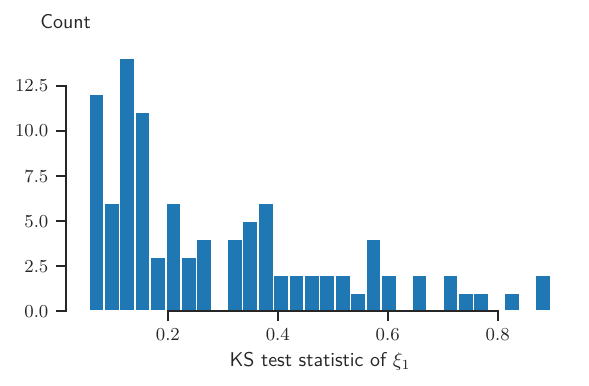}}
     \subfloat{\includegraphics[width=0.45\textwidth]{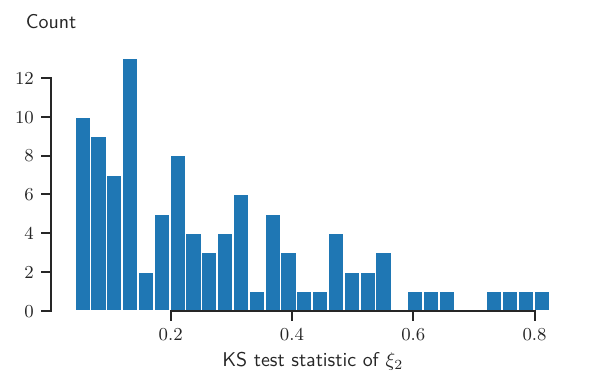}}
     \caption{(Damage location detection) Histograms of KS test statistic values of parameter posteriors for $N_{y} = 100$ samples from the data density.}
    \label{fig:DD_hist}
\end{figure}

We can clearly see that the posterior inferences using our amortization network match the corresponding MCMC estimates and our network is able to infer the center of circular damages accurately conditional on the boundary potential measurements.
This is also reflected in the very low values of re-simulation error ($\mathcal{E}_{\text{re-sim}} = \num{1.06e-2}$) and error metrics based on KS test statistic values in Fig.~\ref{fig:DD_hist} for $N_{y} = 100$ samples from the data density.
%
%
%
%
%
\subsection{Elliptic PDE with uncertain  conductivity field}
\noindent

Consider the 1D steady-state heat conduction equation with no heat sources:
\begin{equation}
-\frac{d}{dx}\left(a(x,\xi)\frac{d}{dx}u(x, \xi)\right) = 0,
\label{eqn: elliptic spde}
\end{equation}
for $x$ in $[0,1]$ and with Dirichlet boundary values:
$$
u(0, \xi) = 1\;\mbox{and}\;u(1, \xi) = 0.
$$
The function $a(x,\xi)$ is a spatially varying conductivity field, and $u(x,\xi)$ is the corresponding temperature field.
We assume that the conductivity is characterized by a random field given by the following analytical equation:
$$
a(x,{\xi}) = \exp\{g(x,{\xi})\},
$$
where
$$
g(x,{\xi}) = \sum_{i=1}^5 \xi_i \frac{\sqrt{2}\sigma}{(i - \frac{1}{2})\pi}\sin\left((i-\frac{1}{2})\pi x\right),
$$
is a random sinusoidal field with uncertain parameters $\xi = \{\xi_i\}_{i=1}^{5}$.
These $\xi_i$'s are independent standard normal random variables with zero mean and unit variance and we consider the variance of the field $\sigma$ to be $1.5$.

Now to demonstrate the effectiveness of our approach, we carry out the problem of inferring the thermal conductivity of this heterogeneous rod based on the observed temperature measurements.
For this, we assume that we have access to a set of $m = 9$ potentially noisy measurements $y_1,y_2, \dots,y_m$ of  $u(x_1,\xi),\dots,u(x_m,\xi)$ at $m$ equidistant points between $0.15$ and $0.85$ along the length of the rod. 
We collectively denote all the noisy measurements as vector $y=(y_1,y_2, \dots,y_m)$.
The inverse problem here is to find the posterior distribution of the uncertain parameters in the conductivity, $p(\xi|y)$, that leads to the observed data $y$.
To do this, we learn the required Bayesian inverse map from $y \in \R^{9}$ to $\xi \in \R^{5}$ using our proposed amortization network \textcolor{black}{for Gaussian guide as in Sec.~\ref{subsubsec:Gaussian guide}}.

To move forward with building our amortization network, we set up the required ingredients likelihood and prior. 
We assume that the $m$ noisy measurements are iid, and  
we model the measurement process using a Gaussian likelihood with a noise scale of $\gamma=0.015$:
$$
{p(y |\xi)} = \prod_{i=1}^{m}N(y_i|u(x_i,\xi),\gamma^2).
$$
To reduce the computational overhead, we construct a neural network approximator of $u(x, \xi)$ as $\hat{u}(x, \xi;\theta)$, using a physics-informed approach that minimizes the energy functional-based loss of Eq.~(\ref{eqn: elliptic spde}) as described in~\cite{karumuri2020simulator}.
Here, $\theta$ are the neural network approximator parameters.
The network takes as inputs a spatial location $x$ and the conductivity parameters $\xi$ and outputs the corresponding temperature $\hat{u}(x, \xi;\theta)$.
This network is a residual network with $5$ residual blocks, each consisting of $3$ layers with $40$ neurons each, and with sigmoid linear units (SiLUs) as the activation function.
Now the likelihood can be approximated as,
\begin{equation}
{p(y |\xi)} \approx \prod_{i=1}^mN(y_i|\hat{u}(x_i, \xi;\theta),\gamma^2),
\end{equation}
where $\hat{u}$ is the forward model. Moving ahead to the prior, we assume that the parameters follow a Gaussian prior, $\xi \sim \mathcal{N}(0, I)$.

Now, for learning the inverse map, we chose the three networks in our amortization network to be feed-forward networks with four hidden layers of sizes $50$, $40$, $30$, and $20$, respectively.
We trained this amortization net following Algorithm~\ref{alg:Inference network training} for a total of $35{,}000$ iterations ($N_{iter}$), with $(N_{y}=64, N_{z}=5)$ samples in each iteration. 
We started with an initial learning rate of $\eta_{I_0}=\num{e-3}$ and used a step decay learning rate schedule with a multiplicative factor of $\alpha=0.5$ after every $r=20{,}000$ iterations.

\begin{figure}[H]
\centering
\subcaptionbox{Posterior conductivity field and the corresponding temperature estimates from our method.\protect\label{fig:EPDE1_AVI}}%
{\includegraphics[width=0.45\textwidth]{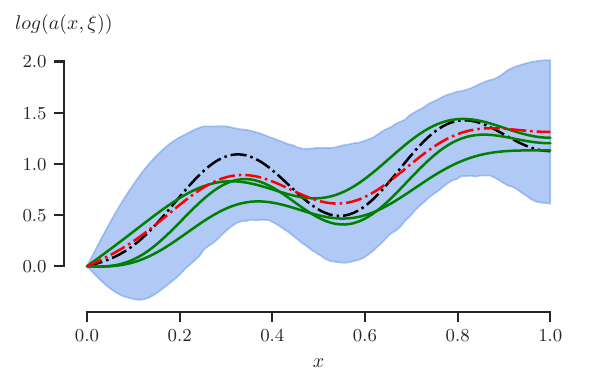}
\includegraphics[width=0.45\textwidth]{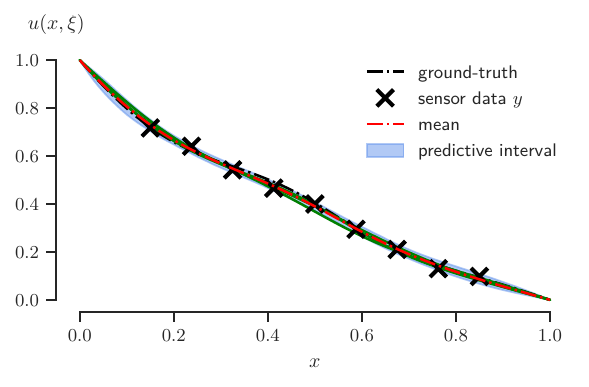}}\\
\subcaptionbox{Posterior conductivity field and the corresponding temperature estimates from MCMC.\protect\label{fig:EPDE1_MCMC}}%
{\includegraphics[width=0.45\textwidth]{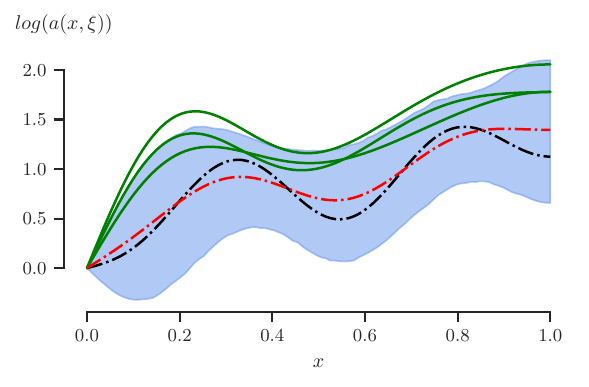}
\includegraphics[width=0.45\textwidth]{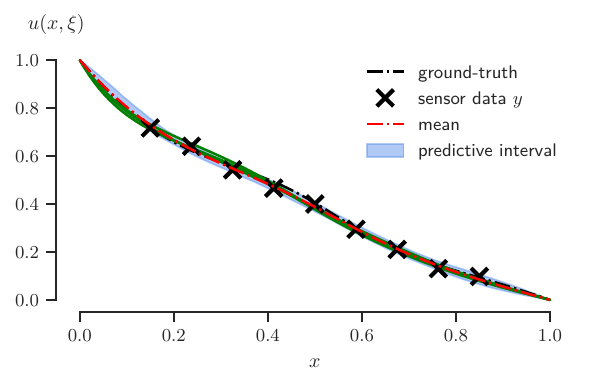}}\\
\subcaptionbox{Pairwise posterior density estimates.\protect\label{fig:EPDE1_pairplot}}%
{\includegraphics[width=0.9\textwidth]{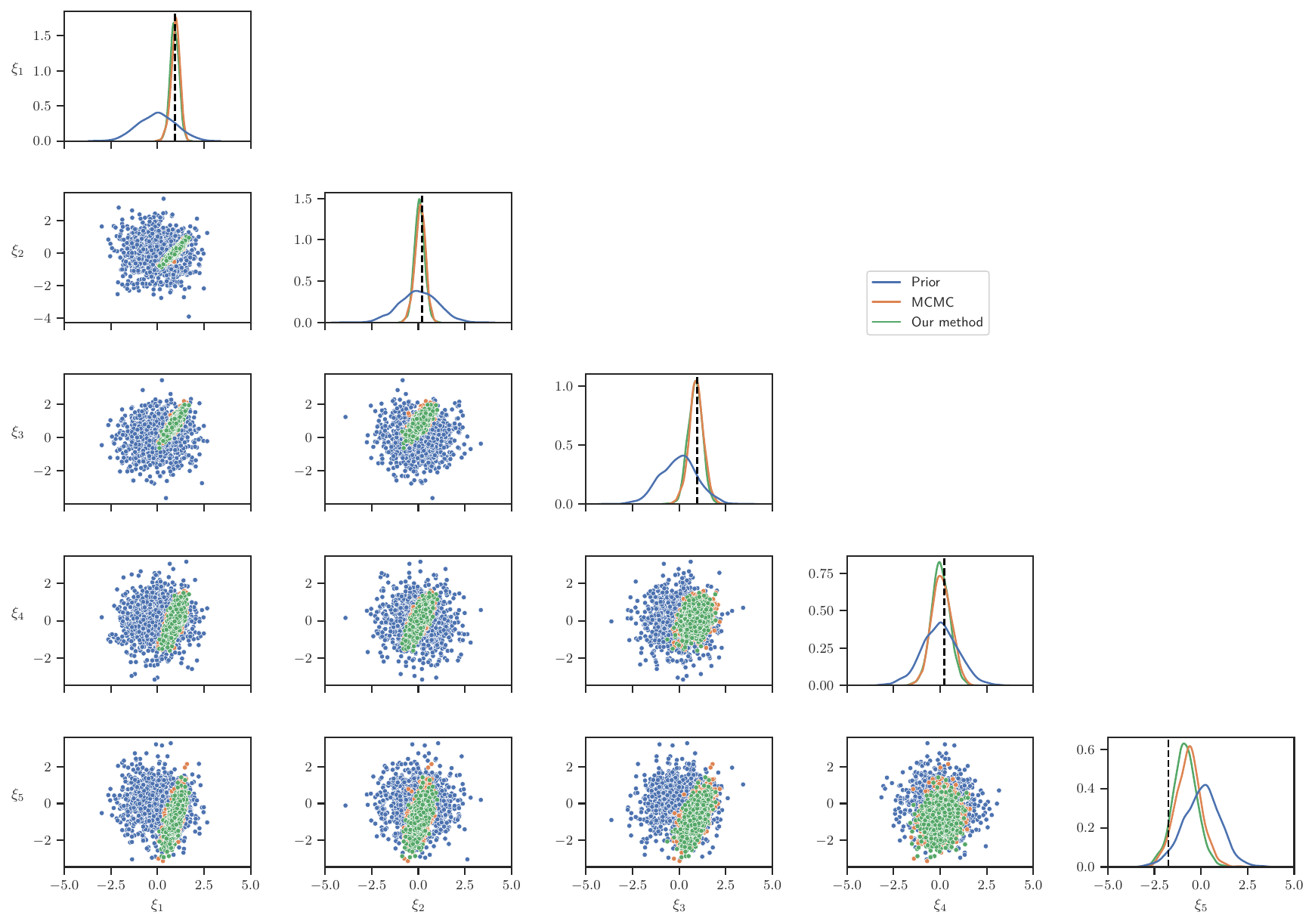}}
\caption{(Elliptic PDE - Observation set 1.)
Qualitative results of the elliptic PDE problem from our method and MCMC approaches.}
\label{fig:EPDE1}
\end{figure}

\begin{figure}[H]
\centering
\subcaptionbox{Posterior conductivity field and the corresponding temperature estimates from our method.\protect\label{fig:EPDE2_AVI}}%
{\includegraphics[width=0.45\textwidth]{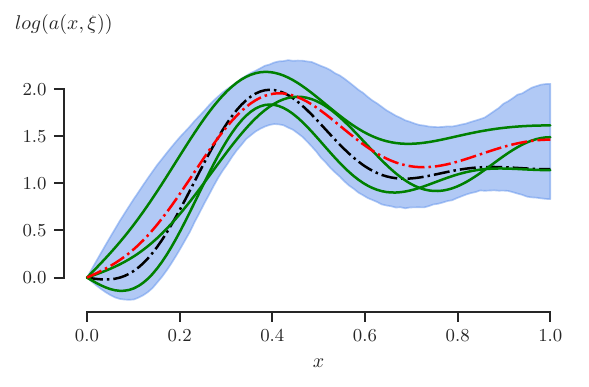}
\includegraphics[width=0.45\textwidth]{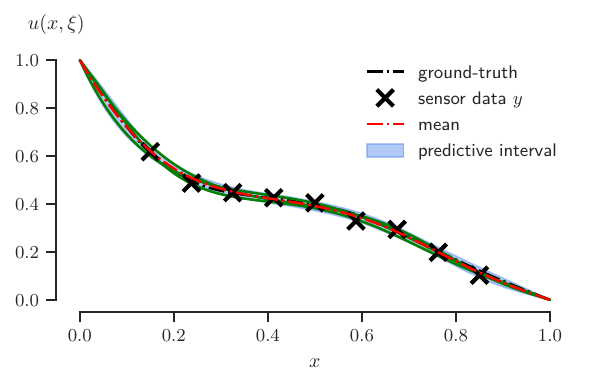}}\\
\subcaptionbox{Posterior conductivity field and the corresponding temperature estimates from MCMC.\protect\label{fig:EPDE2_MCMC}}%
{\includegraphics[width=0.45\textwidth]{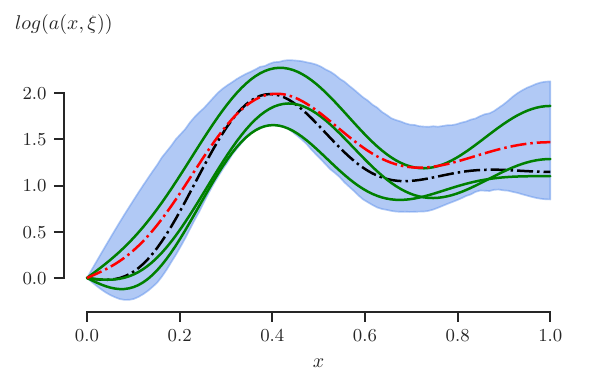}
\includegraphics[width=0.45\textwidth]{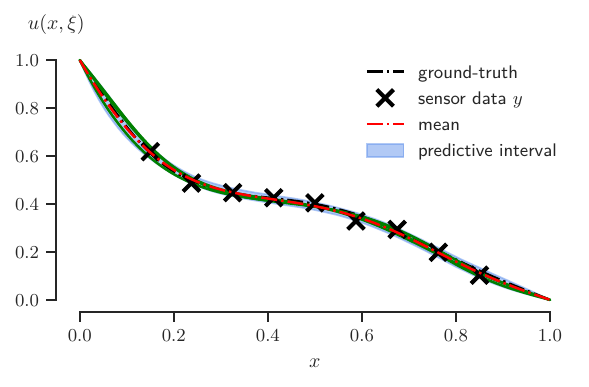}}\\
\subcaptionbox{Pairwise posterior density estimates.\protect\label{fig:EPDE2_pairplot}}%
{\includegraphics[width=0.9\textwidth]{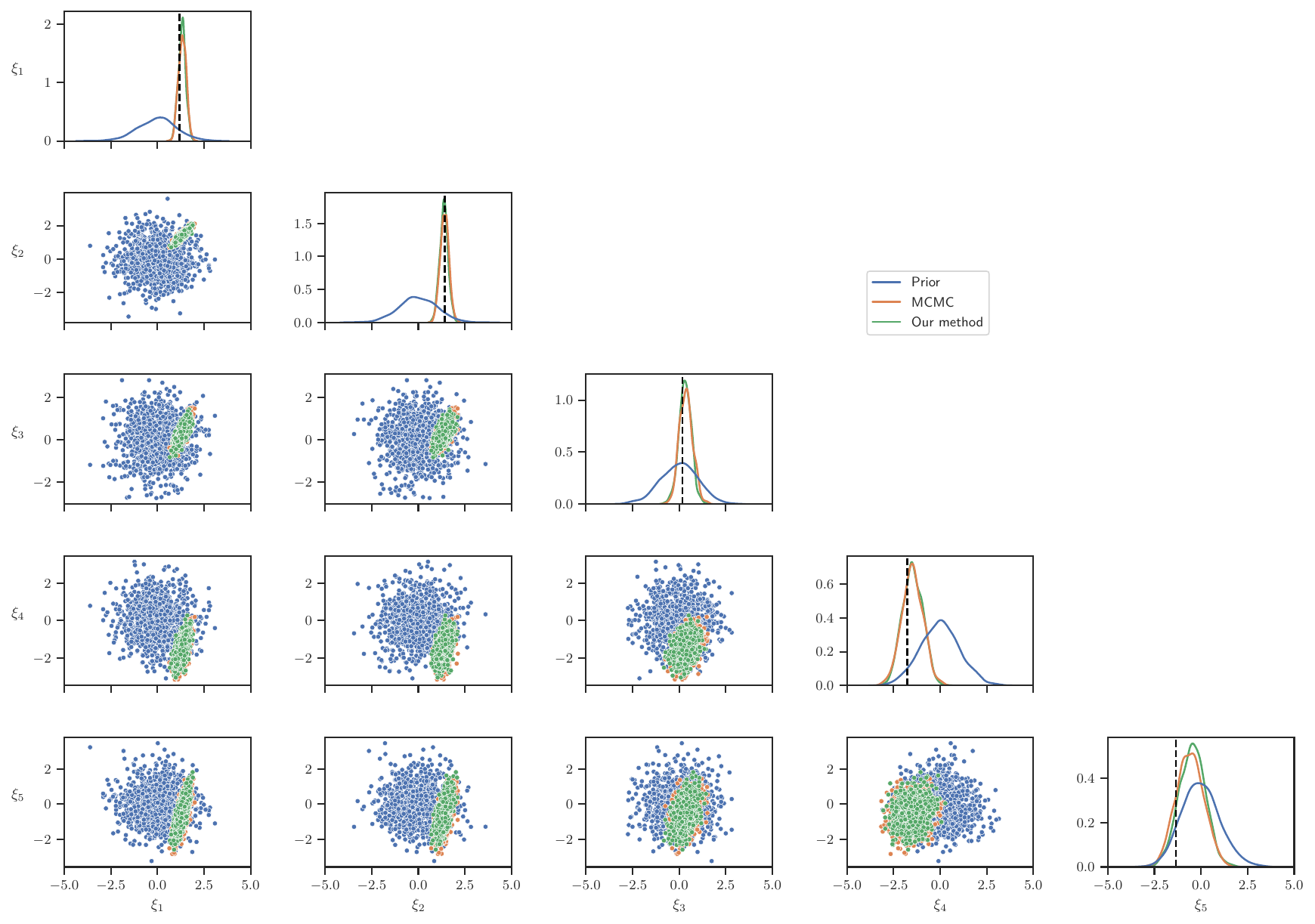}}
\caption{(Elliptic PDE - Observation set 2.)
Qualitative results of the elliptic PDE problem from our method and MCMC approaches.}
\label{fig:EPDE2}
\end{figure}

\begin{figure}[H]
\centering
\subcaptionbox{Posterior conductivity field and the corresponding temperature estimates from our method.\protect\label{fig:EPDE3_AVI}}%
{\includegraphics[width=0.45\textwidth]{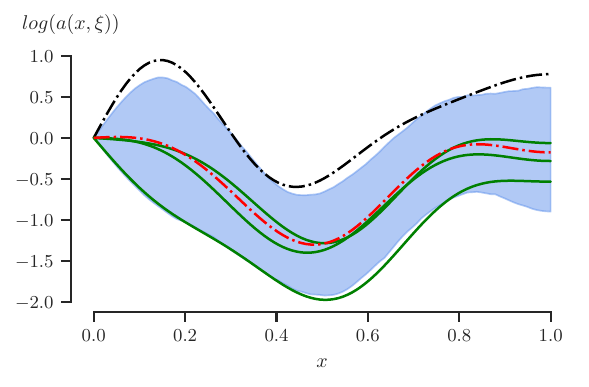}
\includegraphics[width=0.45\textwidth]{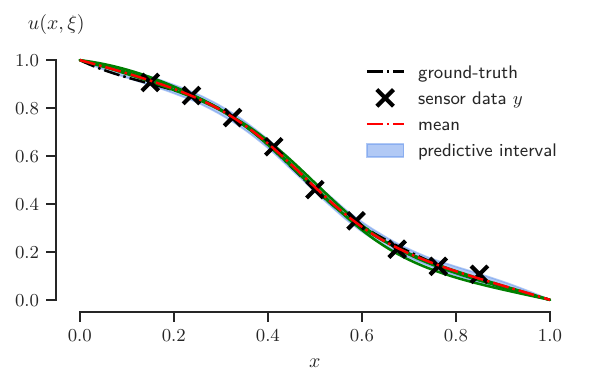}}\\
\subcaptionbox{Posterior conductivity field and the corresponding temperature estimates from MCMC.\protect\label{fig:EPDE3_MCMC}}%
{\includegraphics[width=0.45\textwidth]{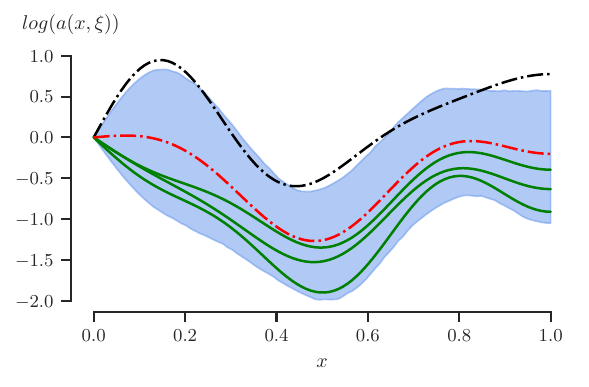}
\includegraphics[width=0.45\textwidth]{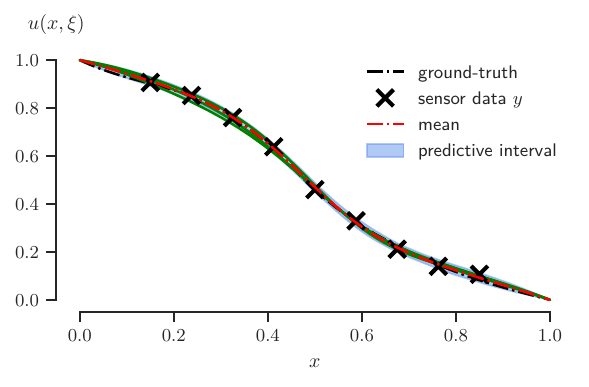}}\\
\subcaptionbox{Pairwise posterior density estimates.\protect\label{fig:EPDE3_pairplot}}%
{\includegraphics[width=0.9\textwidth]{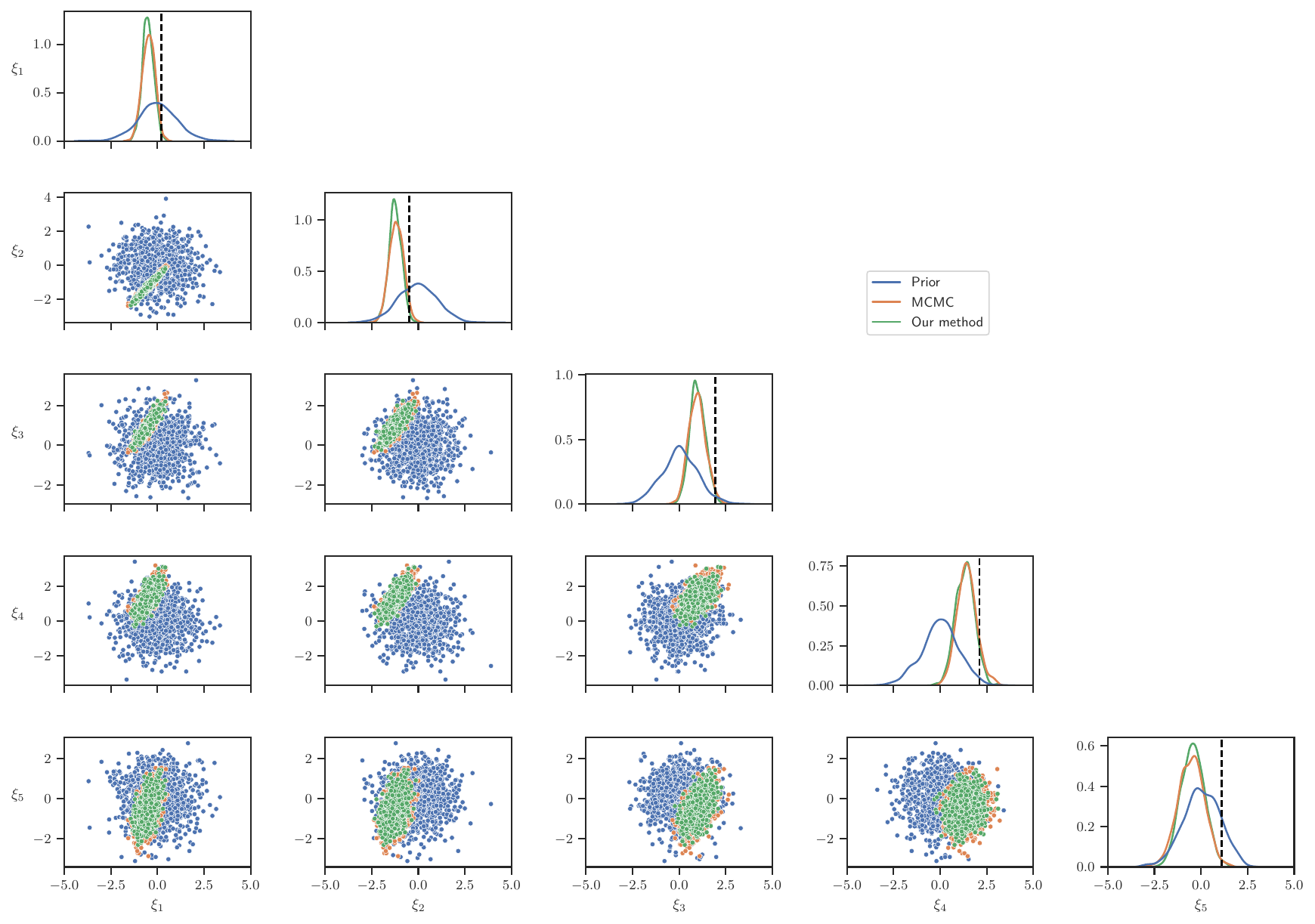}}
\caption{(Elliptic PDE - Observation set 3.)
Qualitative results of the elliptic PDE problem from our method and MCMC approaches.}
\label{fig:EPDE3}
\end{figure}

Qualitative results of the posterior conductivity fields inferred using our amortization network for three sets of observations are shown in Figs.~(\ref{fig:EPDE1} - \ref{fig:EPDE3}), along with corresponding MCMC estimates for comparison. 
The green lines in Figs.~(\ref{fig:EPDE1_AVI}-\ref{fig:EPDE1_MCMC}), (\ref{fig:EPDE2_AVI}-\ref{fig:EPDE2_MCMC}), and, (\ref{fig:EPDE3_AVI}-\ref{fig:EPDE3_MCMC}) represent few samples of the inferred posterior input field and its corresponding solution responses. 
The black dotted line corresponds to the ground truth used to generate the measurement data, and the black crosses mark the measurement data. 
Figs.~\ref{fig:EPDE1_pairplot}, \ref{fig:EPDE2_pairplot} and \ref{fig:EPDE3_pairplot} show the distribution of posterior and prior draws of parameters with a pairplot, where ground truth parameter values are indicated by a black dashed line for reference.

\begin{figure}[H] 
     \centering
     \subfloat{\includegraphics[width=0.45\textwidth]{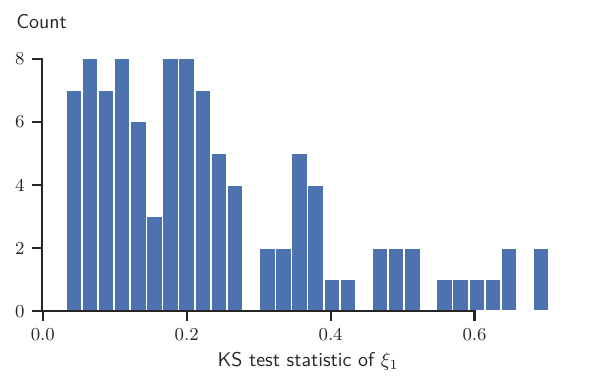}}
     \subfloat{\includegraphics[width=0.45\textwidth]{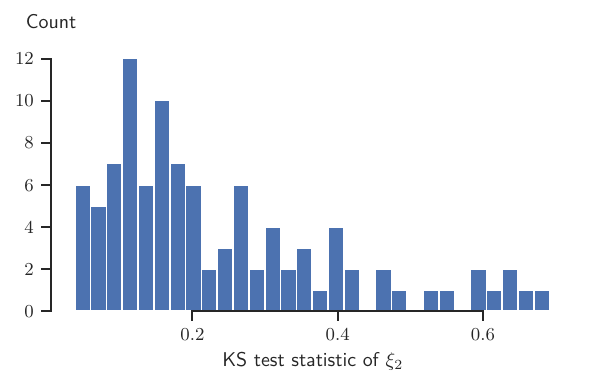}}\\
     \subfloat{\includegraphics[width=0.45\textwidth]{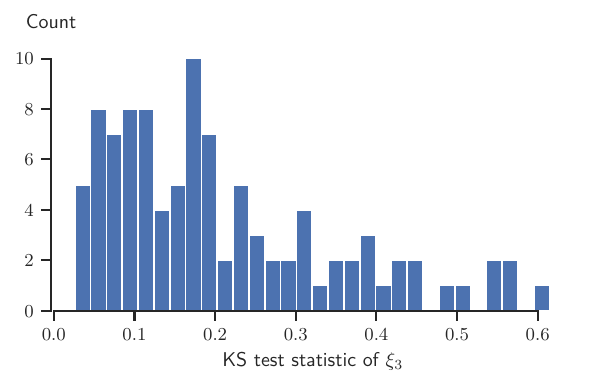}}
     \subfloat{\includegraphics[width=0.45\textwidth]{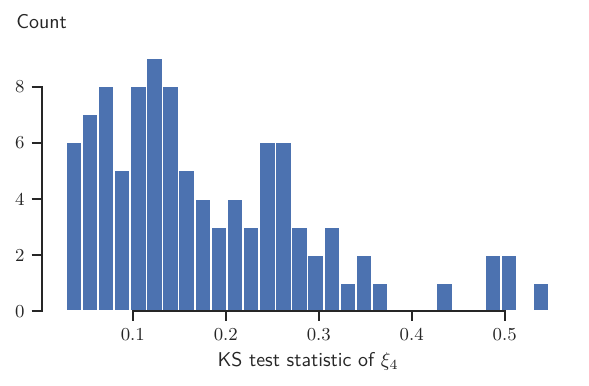}}\\
     \subfloat{\includegraphics[width=0.45\textwidth]{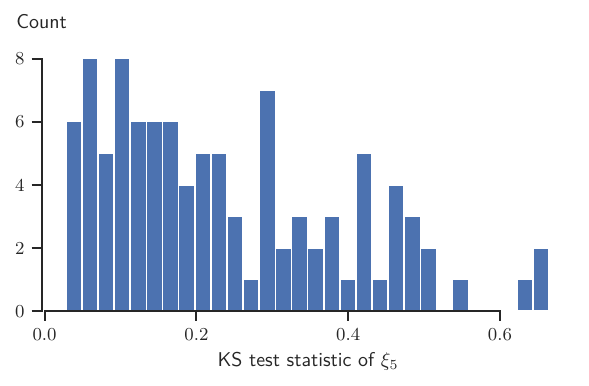}}
     \caption{(Elliptic PDE) Histograms of KS test statistic values of parameter posteriors for $N_{y} = 100$ samples from the data density.}
    \label{fig:EPDE_hist}
\end{figure}

These results demonstrate that the posterior draws of the conductivity field from our amortization network accurately capture a distribution over the true solution, conditional on the noisy observations.
 Moreover, we observe that the ground truth MCMC posterior estimates follow a Multivariate Gaussian distribution and our amortization network can learn it.
Notably, the correlations of the parameters learned from our method are in almost agreement with the corresponding estimates from MCMC.
This is evident from the low values of re-simulation error ($\mathcal{E}_{\text{re-sim}} = \num{4.05e-2}$) and error metrics based on KS test statistic values in Fig.~\ref{fig:EPDE_hist}.
These quantitative results are also estimated using $N_{y} = 100$ samples from the data density.
Overall, these results demonstrate the effectiveness and accuracy of our proposed amortization network for inferring posterior distributions on the fly.
%
%
%
\newpage
%
\newpage
\subsection{Inverse kinematics}
\noindent

\begin{figure}[htb]
  \centering
    \begin{subfigure}[b]{0.47\textwidth}
      \includegraphics[width=\textwidth]{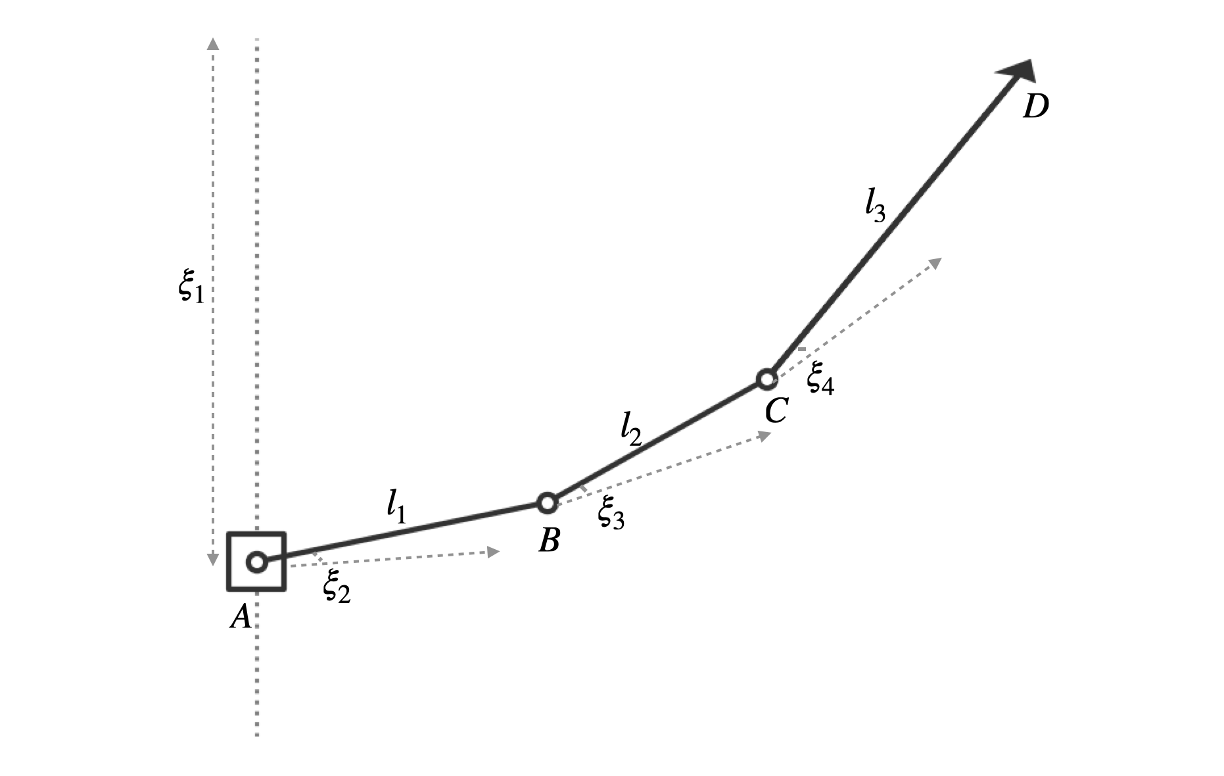}
      \caption{}
      \label{fig:schematic_inverse_kinematics}
    \end{subfigure}
    \begin{subfigure}[b]{0.47\textwidth}
      \includegraphics[width=\textwidth]{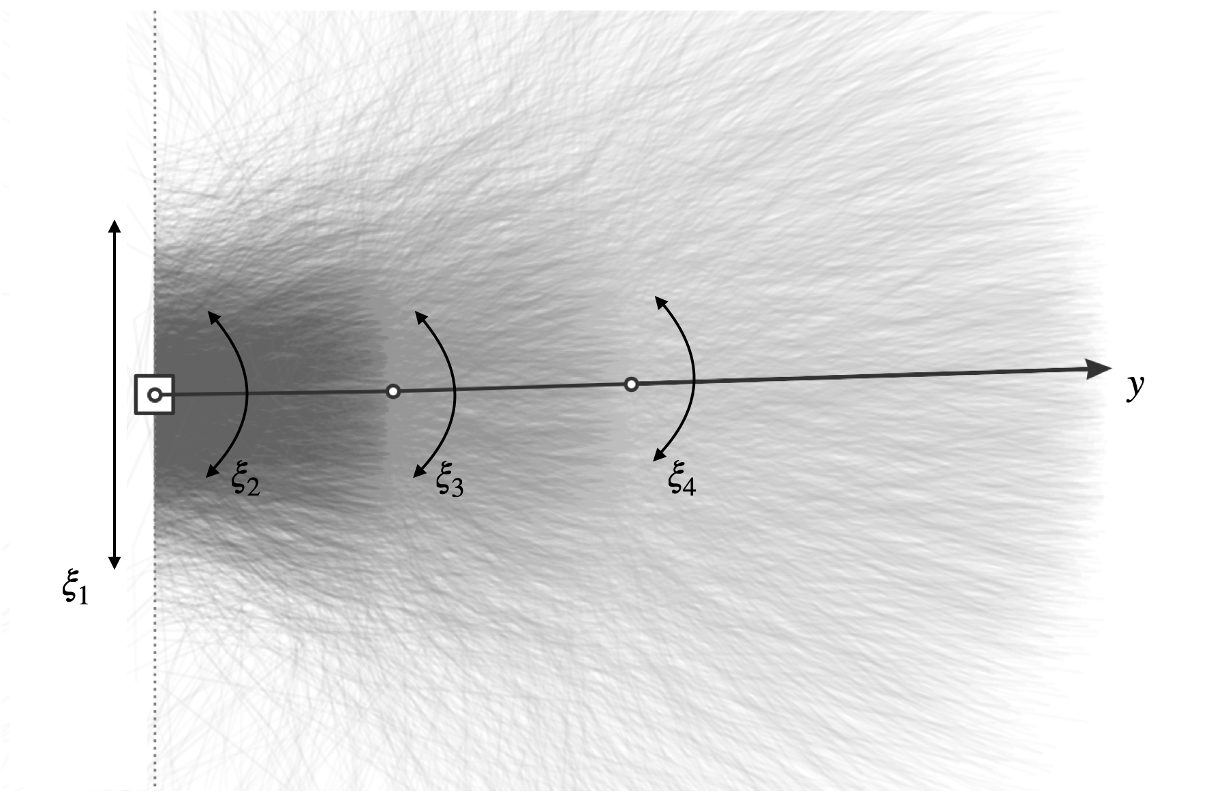}
      \caption{}
      \label{fig:prior_inverse_kinematics}
    \end{subfigure}
  \caption{(Inverse Kinematics) Illustration~(a) of the articulated arm with three segments mounted on a rail with a slider and~(b) prior distribution of the parameters $\xi$ visualized as a collection of possible arm positions.}
  \label{fig:1dim_eg_histograms}
  \end{figure}

We consider the inverse kinematics problem of identifying the configuration of a multi-jointed $2D$ arm that ends at a given position, see Fig.~\ref{fig:schematic_inverse_kinematics}.
This problem has been considered in~\cite{ardizzone2018analyzing}.

The forward model takes the height on the slider $\xi_1$ and the three joint angles $\xi_2, \xi_3, \xi_4,$ and returns the coordinates of the arm end point $f(\xi) = (f_1(\xi), f_2(\xi))$:
$$f_1(\xi) = l_1 \cos(\xi_2)+l_2 \cos(\xi_2+\xi_3)+l_3 \cos(\xi_2+\xi_3+\xi_4),$$
$$f_2(\xi) = \xi_1+l_1 \sin(\xi_2)+l_2 \sin(\xi_2+\xi_3)+l_3 \sin(\xi_2+\xi_3+\xi_4),$$
with arm lengths $l_1=0.5, l_2=0.5$ and $l_3=1.$
The parameters $\xi$ follow a Gaussian prior $\xi \sim \mathcal{N}(0, \operatorname{diag}(\sigma^2))$ with $\sigma= \left(\frac{1}{4}, \frac{1}{2}, \frac{1}{2}, \frac{1}{2}\right)$ (see Fig.~\ref{fig:prior_inverse_kinematics}).
We assume that we have access to a noisy version $y$ of the arm end coordinates $f(\xi)$.
The likelihood of observed data is chosen to be Gaussian,
$$
p(y|\xi) = \mathcal{N}(y|f(\xi), \gamma^2I),
$$
with $\gamma=0.01$.
The inverse problem is to find the posterior distribution $p(\xi|y)$ of all arm configurations $\xi$ that end at the observed $2D$ position $y$.

\textcolor{black}{To begin with, first we learned the required Bayesian inverse map from $y$ in $\R^2$ to $\xi$ in $\R^4$ using our amortized Gaussian guide as detailed in Sec.~\ref{subsubsec:Gaussian guide}}.
Similar to the before examples, the three distinct neural networks in our amortization network are feed-forward networks with two hidden layers of sizes $20$ and $10$ respectively.
The amortization network was trained following the process described in Algorithm~\ref{alg:Inference network training} for $10{,}000$ iterations ($N_{\text{iter}}$) using  $N_{y}=32$ and $N_{z} =5$ samples in each iteration. 
A step decay learning rate schedule was employed to optimize this network. 
The learning rate was initialized to an initial value of $\eta_{I_0}=\num{e-2}$ and decreased by a multiplicative factor of $\alpha=0.1$ after every $r=5{,}000$ iterations.

\begin{figure}[H]
\centering
\subcaptionbox{Posterior arm configurations from our method.\protect\label{fig:IK-G1_AVI}}%
{\includegraphics[width=0.45\textwidth]{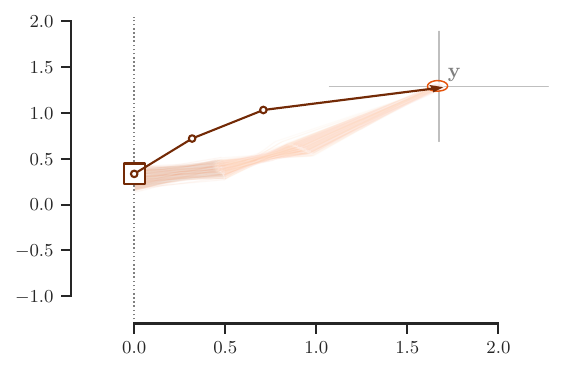}}
\subcaptionbox{Posterior arm configurations from MCMC.\protect\label{fig:IK-G1_MCMC}}%
{\includegraphics[width=0.45\textwidth]{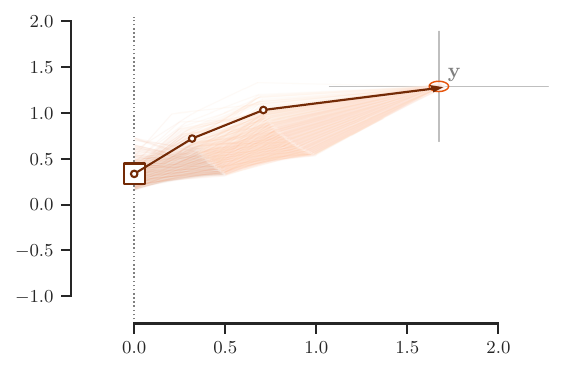}}\\
\subcaptionbox{Pairwise posterior density estimates.\protect\label{fig:IK-G1_pairplot}}%
{\includegraphics[width=0.9\textwidth]{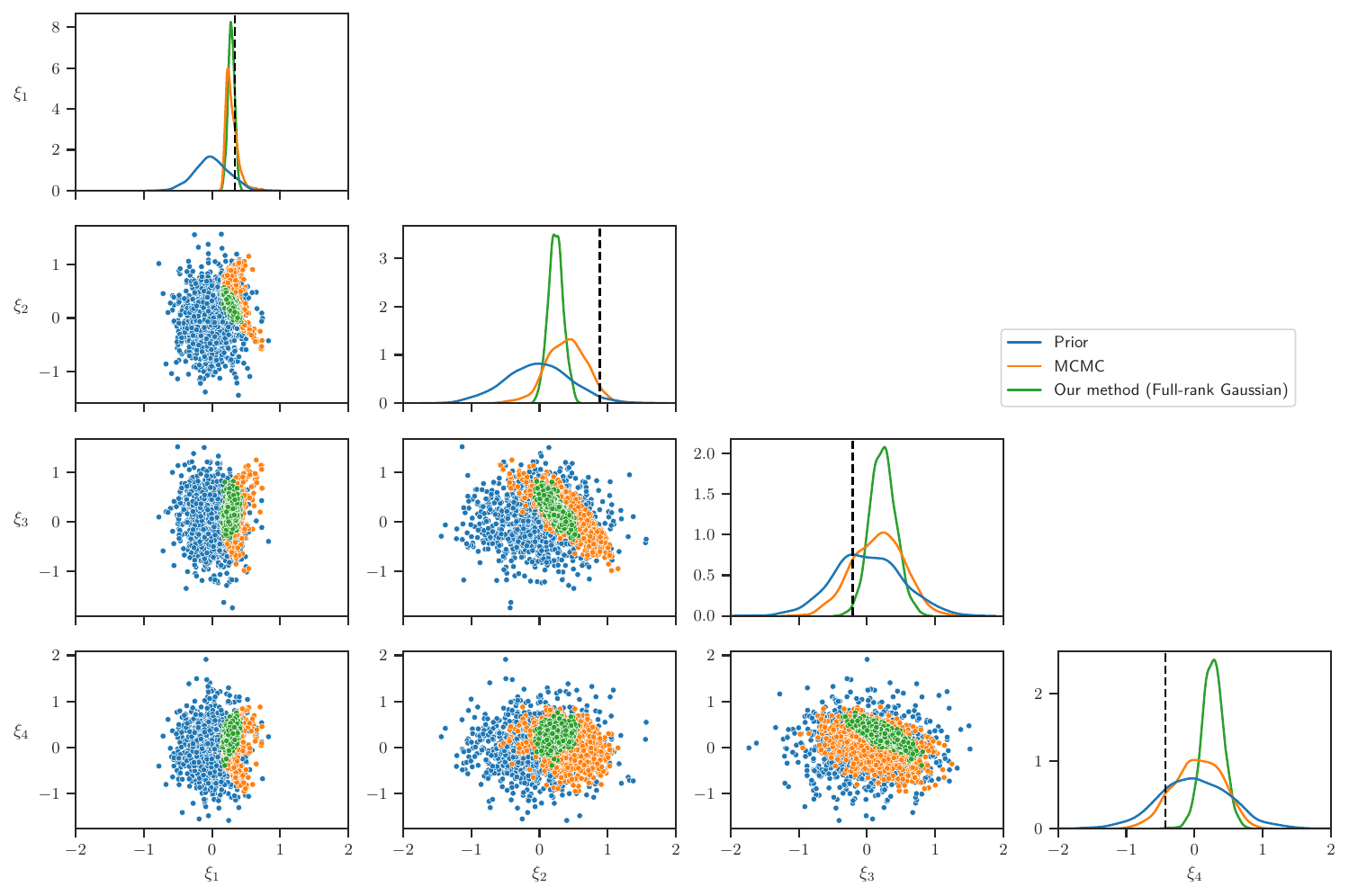}}
\caption{(Inverse Kinematics - Observation set 1.)
Qualitative results of the inverse kinematic problem from Our method \textcolor{black}{(full-rank Gaussian)} and MCMC approaches for the case where the arm ends at a position $y=(1.67, 1.29)$.}
\label{fig:IK-G1}
\end{figure}

\begin{figure}[H]
\centering
\subcaptionbox{Posterior arm configurations from our method.\protect\label{fig:IK-G2_AVI}}%
{\includegraphics[width=0.45\textwidth]{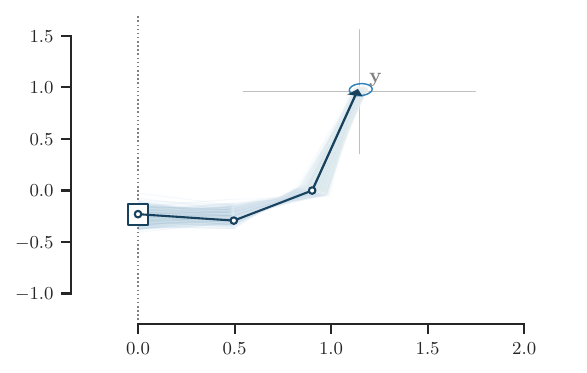}}
\subcaptionbox{Posterior arm configurations from MCMC.\protect\label{fig:IK-G2_MCMC}}%
{\includegraphics[width=0.45\textwidth]{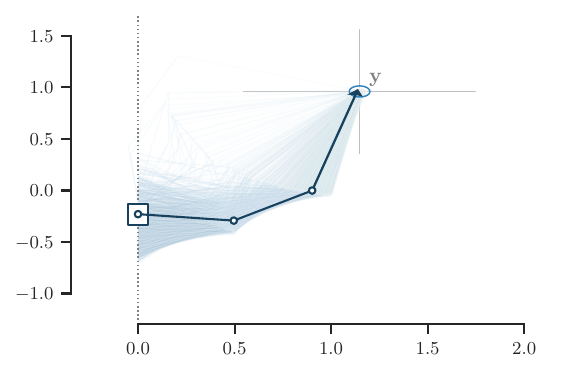}}\\
\subcaptionbox{Pairwise posterior density estimates.\protect\label{fig:IK-G2_pairplot}}%
{\includegraphics[width=0.9\textwidth]{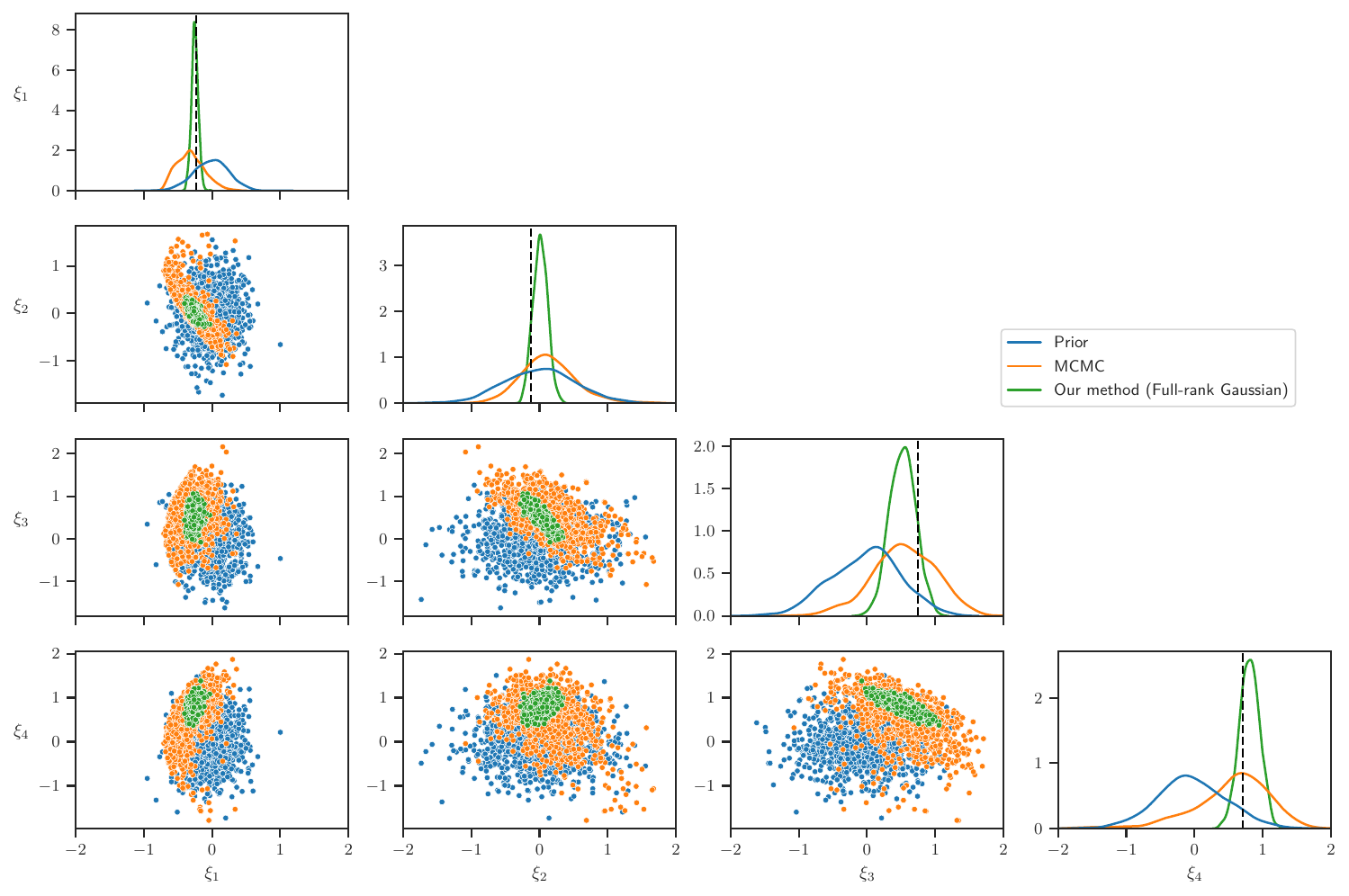}}
\caption{(Inverse Kinematics - Observation set 2.)
Qualitative results of the inverse kinematic problem from Our method \textcolor{black}{(full-rank Gaussian)} and MCMC approaches for the case where the arm ends at a position $y=(1.15, 0.96)$.}
\label{fig:IK-G2}
\end{figure}

\begin{figure}[H]
\centering
\subcaptionbox{Posterior arm configurations from our method.\protect\label{fig:IK-G3_AVI}}%
{\includegraphics[width=0.45\textwidth]{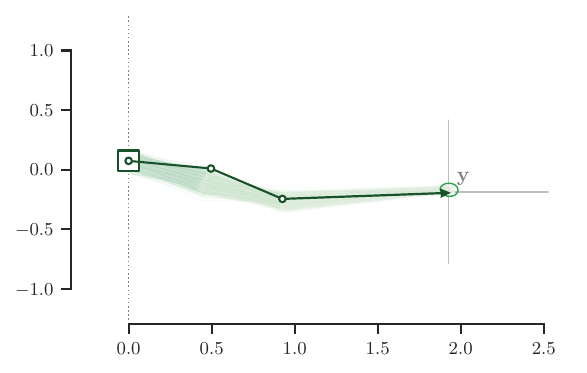}}
\subcaptionbox{Posterior arm configurations from MCMC.\protect\label{fig:IK-G3_MCMC}}%
{\includegraphics[width=0.45\textwidth]{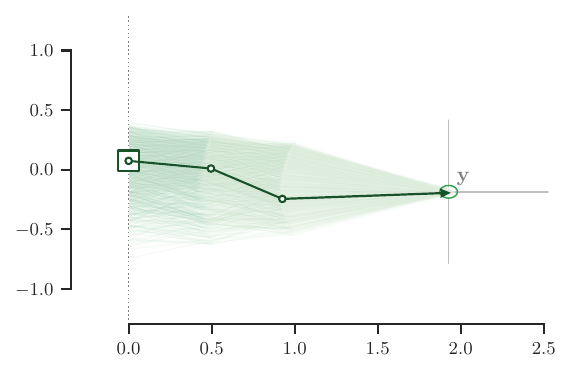}}\\
\subcaptionbox{Pairwise posterior density estimates.\protect\label{fig:IK-G3_pairplot}}%
{\includegraphics[width=0.9\textwidth]{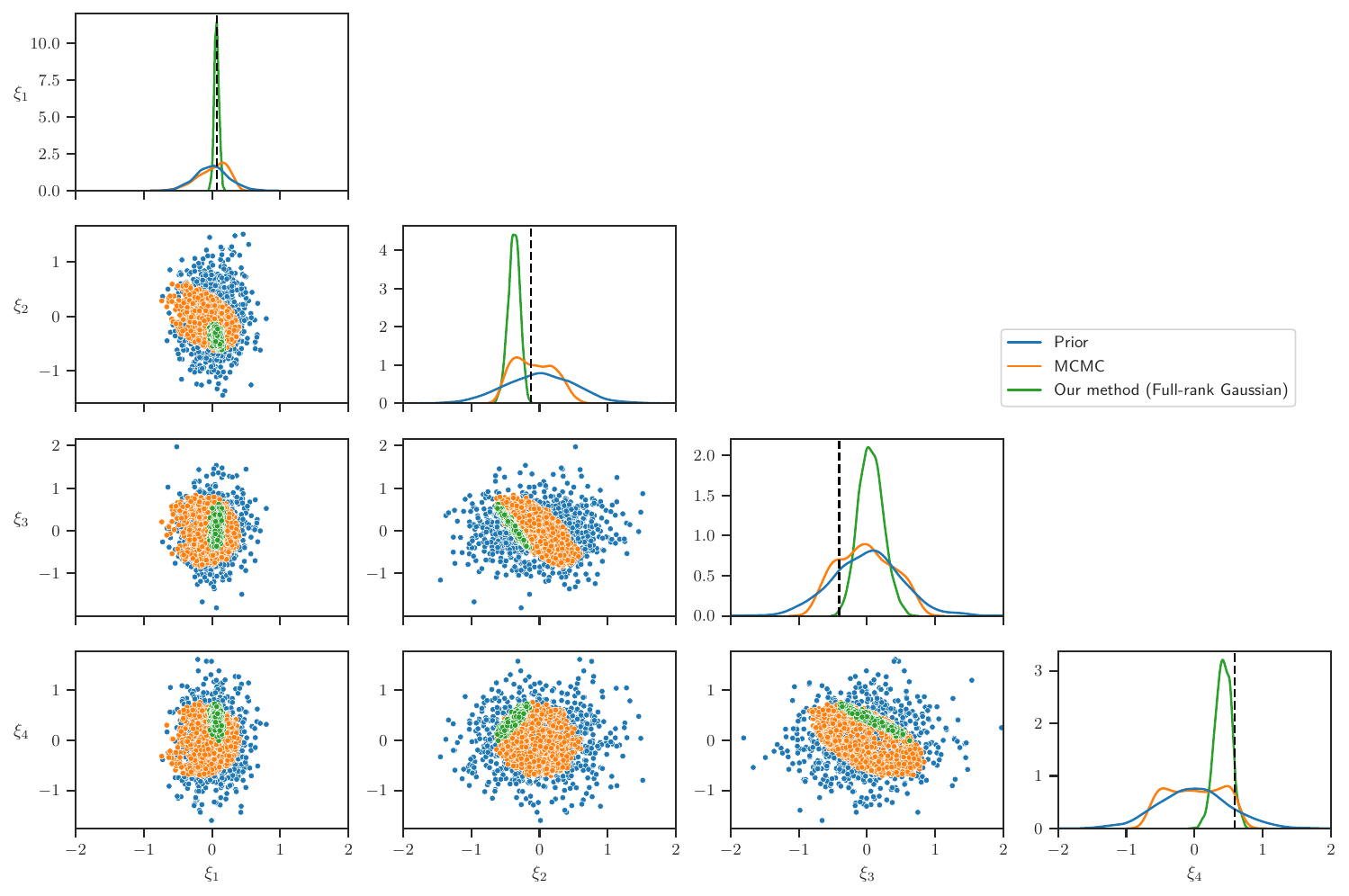}}
\caption{(Inverse Kinematics - Observation set 3.)
Qualitative results of the inverse kinematic problem from Our method \textcolor{black}{(full-rank Gaussian)} and MCMC approaches for the case where the arm ends at a position $y=(1.93, -0.18)$.}
\label{fig:IK-G3}
\end{figure}

\begin{figure}[H]
\centering
\subcaptionbox{Posterior arm configurations from our method.\protect\label{fig:IK-G4_AVI}}%
{\includegraphics[width=0.45\textwidth]{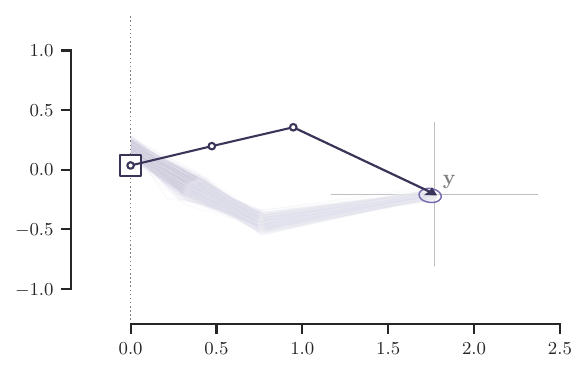}}
\subcaptionbox{Posterior arm configurations from MCMC.\protect\label{fig:IK-G4_MCMC}}%
{\includegraphics[width=0.45\textwidth]{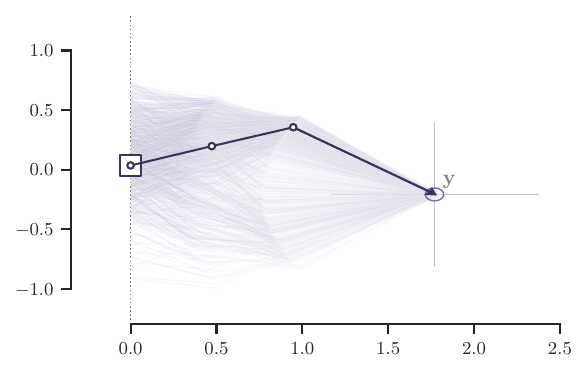}}\\
\subcaptionbox{Pairwise posterior density estimates.\protect\label{fig:IK-G4_pairplot}}%
{\includegraphics[width=0.9\textwidth]{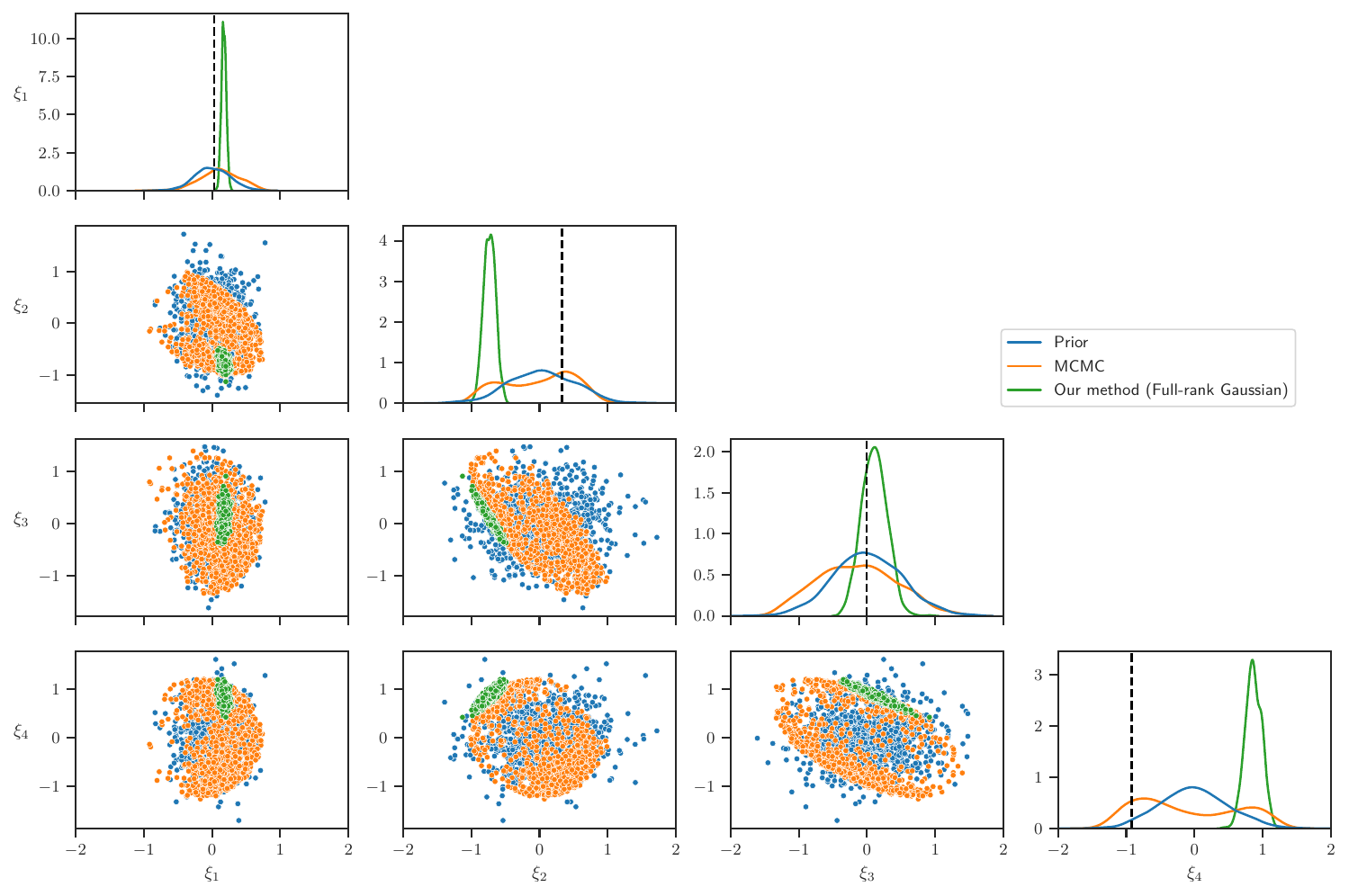}}
\caption{(Inverse Kinematics - Observation set 4.)
Qualitative results of the inverse kinematic problem from Our method \textcolor{black}{(full-rank Gaussian)} and MCMC approaches for the case where the arm ends at a position $y=(1.77, -0.21)$.}
\label{fig:IK-G4}
\end{figure}

\begin{figure}[H]
\centering
\subcaptionbox{Posterior arm configurations from our method.\protect\label{fig:IK-G5_AVI}}%
{\includegraphics[width=0.45\textwidth]{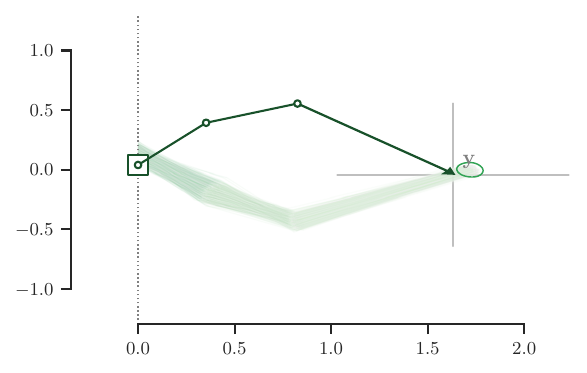}}
\subcaptionbox{Posterior arm configurations from MCMC.\protect\label{fig:IK-G5_MCMC}}%
{\includegraphics[width=0.45\textwidth]{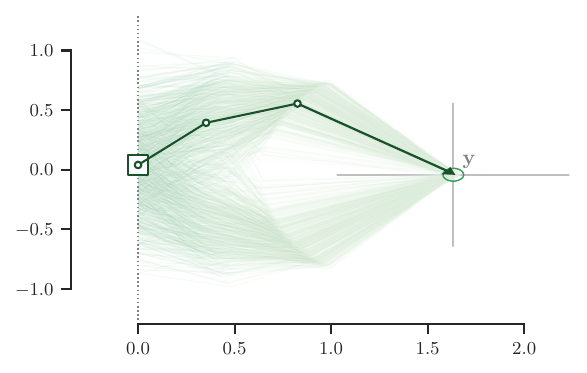}}\\
\subcaptionbox{Pairwise posterior density estimates.\protect\label{fig:IK-G5_pairplot}}%
{\includegraphics[width=0.9\textwidth]{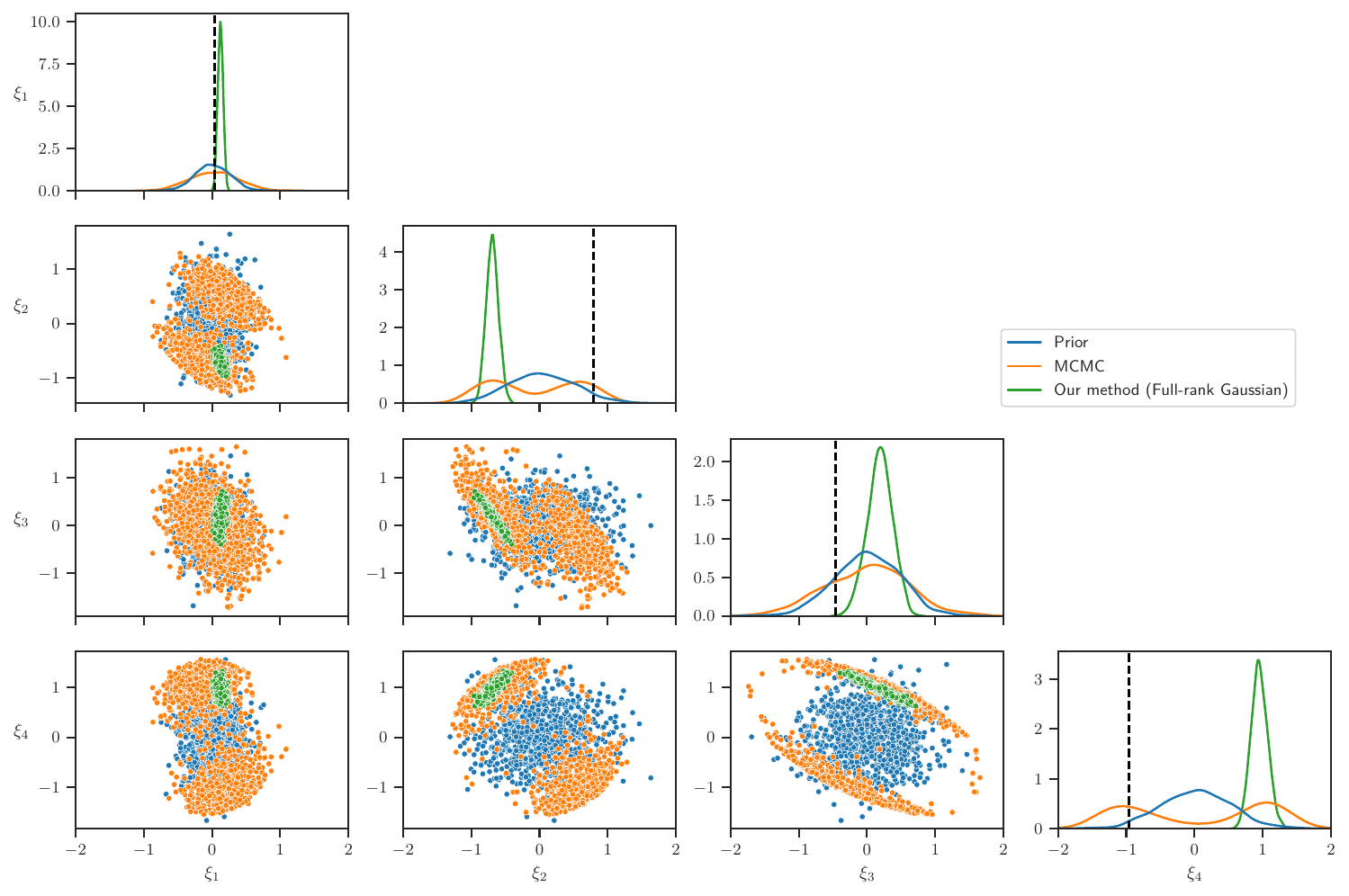}}
\caption{(Inverse Kinematics - Observation set 5.)
Qualitative results of the inverse kinematic problem from Our method \textcolor{black}{(full-rank Gaussian)} and MCMC approaches for the case where the arm ends at a position $y=(1.63, -0.04)$.}
\label{fig:IK-G5}
\end{figure}

Qualitative results of the posteriors learned using our amortization network for five endpoint observation $y$ cases are shown in Figs.~(\ref{fig:IK-G1}-\ref{fig:IK-G5}), along with comparisons against the corresponding MCMC estimates.
Specifically, Figs.~(\ref{fig:IK-G1_AVI}-\ref{fig:IK-G1_MCMC}), (\ref{fig:IK-G2_AVI}-\ref{fig:IK-G2_MCMC}), (\ref{fig:IK-G3_AVI}-\ref{fig:IK-G3_MCMC}), (\ref{fig:IK-G4_AVI}-\ref{fig:IK-G4_MCMC}), (\ref{fig:IK-G5_AVI}-\ref{fig:IK-G5_MCMC}) show the distribution of arm configurations, conditional on the endpoint $y$ marked by a grey cross. 
Here the vertical dotted line represents the rail the arm is based on, and the solid line represents the ground truth arm configuration. 
The faint-colored lines are sampled posterior arm configurations and contour lines around the target represent the area containing $97\%$ of the sampled arm endpoints. 
Figs.~\ref{fig:IK-G1_pairplot}, \ref{fig:IK-G2_pairplot}, \ref{fig:IK-G3_pairplot}, \ref{fig:IK-G4_pairplot} and \ref{fig:IK-G5_pairplot} show the pairplot of parameters from our approach with \textcolor{black}{amortized Gaussian guide} and MCMC.
The diagonal elements show the marginal posterior estimates of the parameters and the off-diagonal elements show the scatter plot for each pair of parameters. 
Ground truth parameter values are marked by a black dashed line for reference on the diagonal elements.

\begin{figure}[H]
  \centering
  \subfloat{\includegraphics[width=0.45\textwidth]{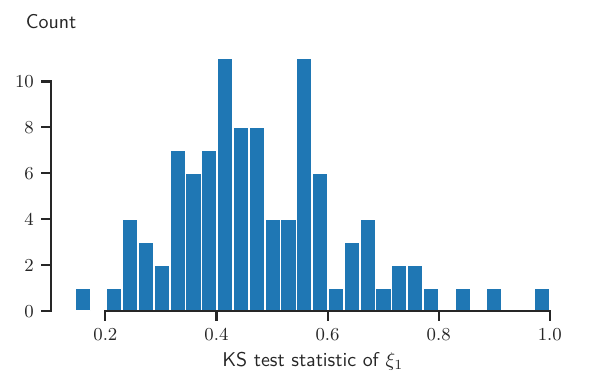}}
  \subfloat{\includegraphics[width=0.45\textwidth]{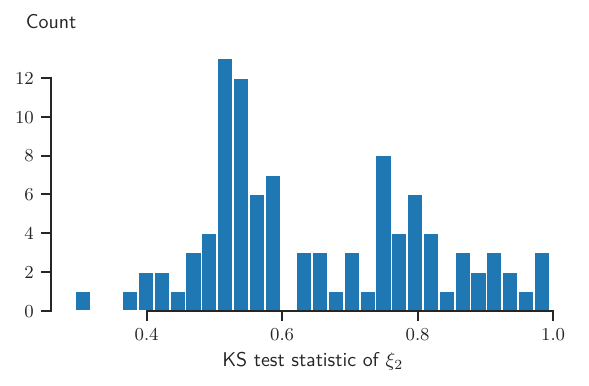}}\\
  \subfloat{\includegraphics[width=0.45\textwidth]{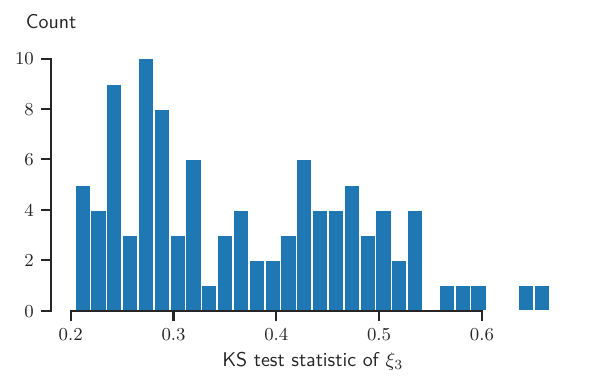}}
  \subfloat{\includegraphics[width=0.45\textwidth]{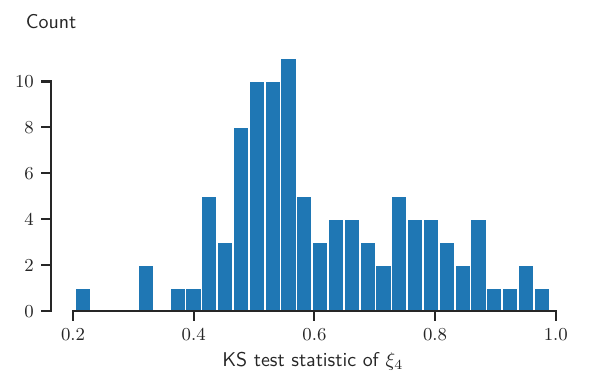}}
  \caption{(Inverse Kinematics) Histograms of KS test statistic values of parameter posteriors \textcolor{black}{with amortized Gaussian guide} for $N_{y} = 100$ samples from the data density.}
 \label{fig:IK-G_hist}
\end{figure}

From these, we see that \textcolor{black}{our amortized Gaussian guide} is able to capture a valid set of arm configurations but not all possible configurations as expected owing to the fact that our chosen posterior guide is a multivariate Gaussian whereas the ground-truth posterior is non-Gaussian and multi-modal.
This is reflected in the low value of re-simulation error ($\mathcal{E}_{\text{re-sim}} = \num{2.32e-2}$) and high values of error metrics based on KS test statistic values in Fig.~\ref{fig:IK-G_hist}.
These quantitative results are estimated using $N_{y} = 100$ samples from the data density.
\textcolor{black}{However, from the pairplots depicted in Figs.~(\ref{fig:IK-G1}-\ref{fig:IK-G5}), we can clearly see that our Gaussian posteriors are effectively capturing the mode of the true posterior (MCMC) as expected. 
This outcome is happening because our approach to learning the amortization network parameters relies on maximizing the AELBO. 
Notably, maximizing AELBO is equivalent to minimizing the KL divergence between approximate and true posterior for all possible datasets.
The inherent nature of this as a mode-seeking approach substantiates our observed results.}
The posterior estimates, in this case, could be improved to capture the complete uncertainty by choosing a variational posterior guide that reflects the non-Gaussian nature and multi-modalities.

\textcolor{black}{To address this need, we employed a conditional normalizing flow guide, as elaborated in Section~\ref{subsubsec:Conditional normalizing flow guide}, to effectively learn these complex posteriors.
The conditional normalizing flow model was configured with a length of $K=15$ incorporating a CCB followed by a fixed random permutation in each step of the flow.
The internal networks, denoted as `$s$' and `$a$' within each CCB, are considered to be feed-forward networks with two hidden layers of size 100 each, leaky ReLU activations between the layers, and a tanh activation at the output.
Similarly, the inner networks `$t$' and `$b$' within each CCB are feed-forward networks with two hidden layers of size 100 with leaky ReLU activation and a linear activation at the output. 
The parameters of this flow model (aka amortization network) were trained following the process described in Algorithm~\ref{alg:Inference network training} for $ 200{,}000 $ iterations ($N_{\text{iter}}$) using  $N_{y}=64$ and $N_{z} =32$ samples in each iteration. 
For the optimization process, Adam optimizer with a constant learning rate of $\num{e-3}$ across all iterations was used with AMSGrad~\cite{reddi2019convergence}, a new exponential moving average variant to improve the convergence.}

\begin{figure}[H]
\centering
\subcaptionbox{Posterior arm configurations from our method.\protect\label{fig:IK-CNF1_AVI}}%
{\includegraphics[width=0.45\textwidth]{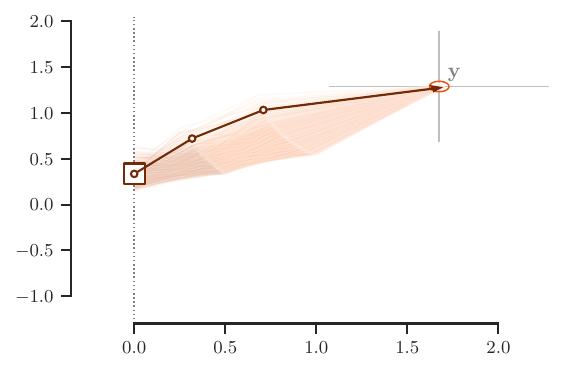}}
\subcaptionbox{Posterior arm configurations from MCMC.\protect\label{fig:IK-CNF1_MCMC}}%
{\includegraphics[width=0.45\textwidth]{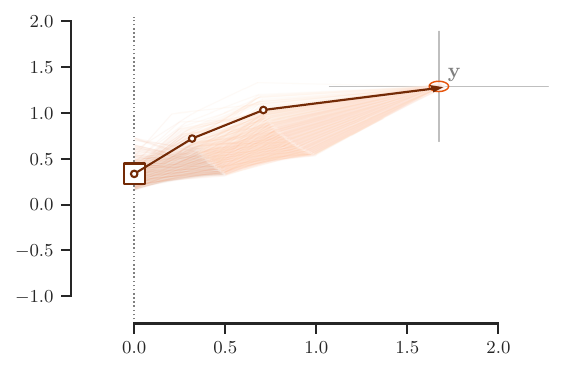}}\\
\subcaptionbox{Pairwise posterior density estimates.\protect\label{fig:IK-CNF1_pairplot}}%
{\includegraphics[width=0.9\textwidth]{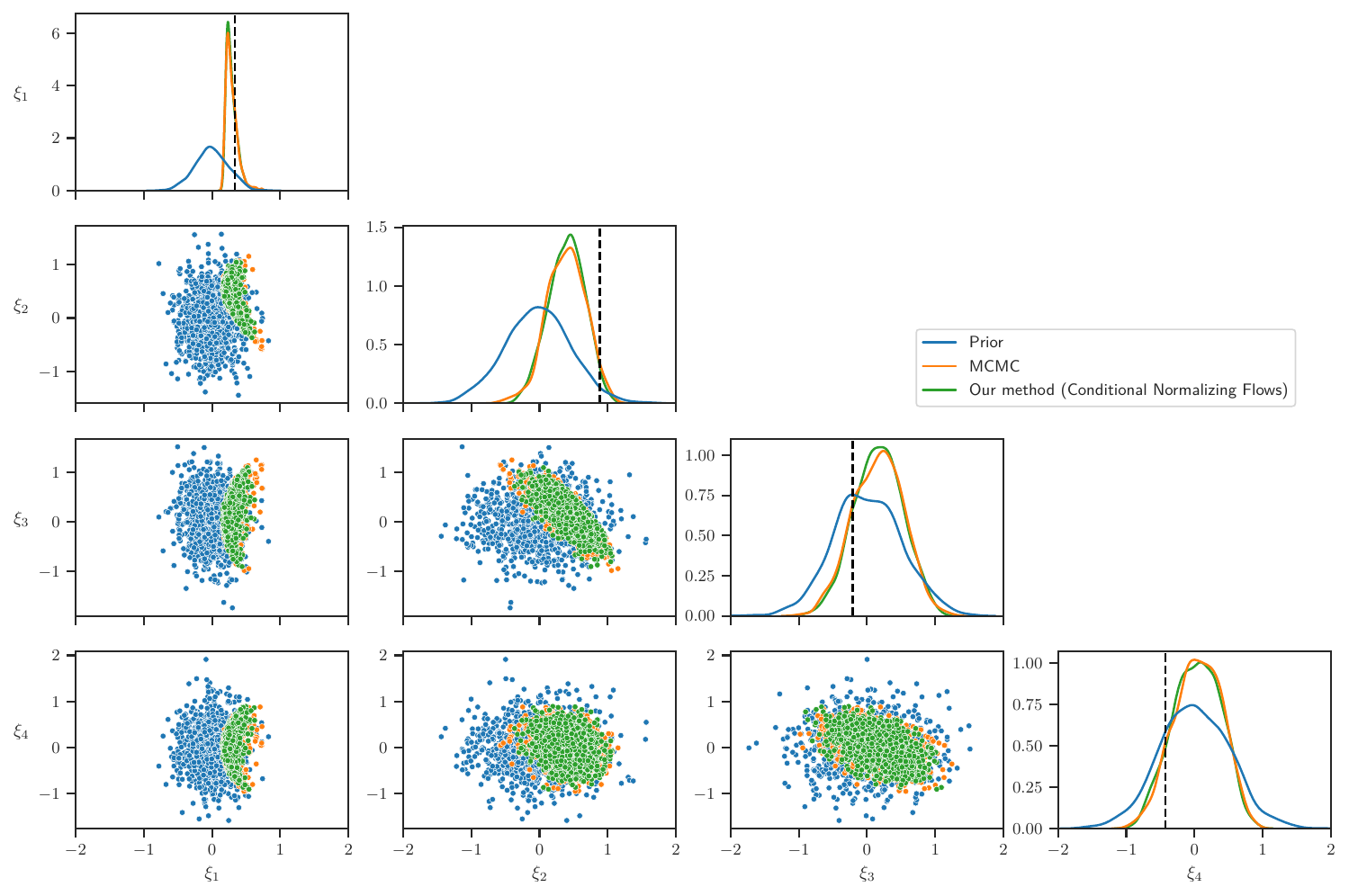}}
\caption{\textcolor{black}{(Inverse Kinematics - Observation set 1.)
Qualitative results of the inverse kinematic problem from Our method (Conditional normalizing flow) and MCMC approaches for the case where the arm ends at a position $y=(1.67, 1.29)$.}}
\label{fig:IK-CNF1}
\end{figure}

\begin{figure}[H]
\centering
\subcaptionbox{Posterior arm configurations from our method.\protect\label{fig:IK-CNF2_AVI}}%
{\includegraphics[width=0.45\textwidth]{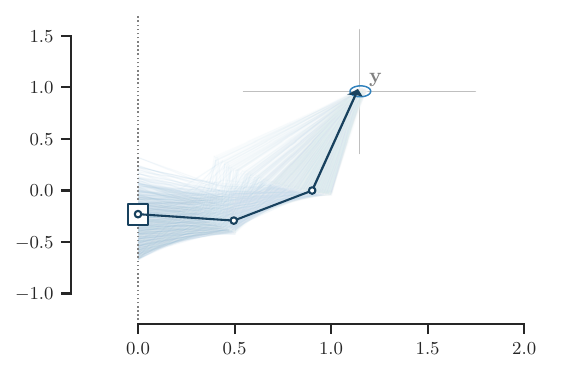}}
\subcaptionbox{Posterior arm configurations from MCMC.\protect\label{fig:IK-CNF2_MCMC}}%
{\includegraphics[width=0.45\textwidth]{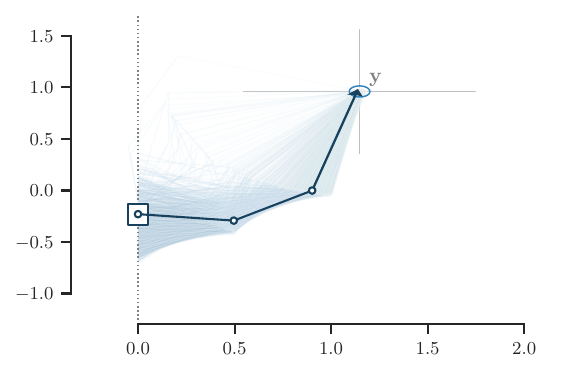}}\\
\subcaptionbox{Pairwise posterior density estimates.\protect\label{fig:IK-CNF2_pairplot}}%
{\includegraphics[width=0.9\textwidth]{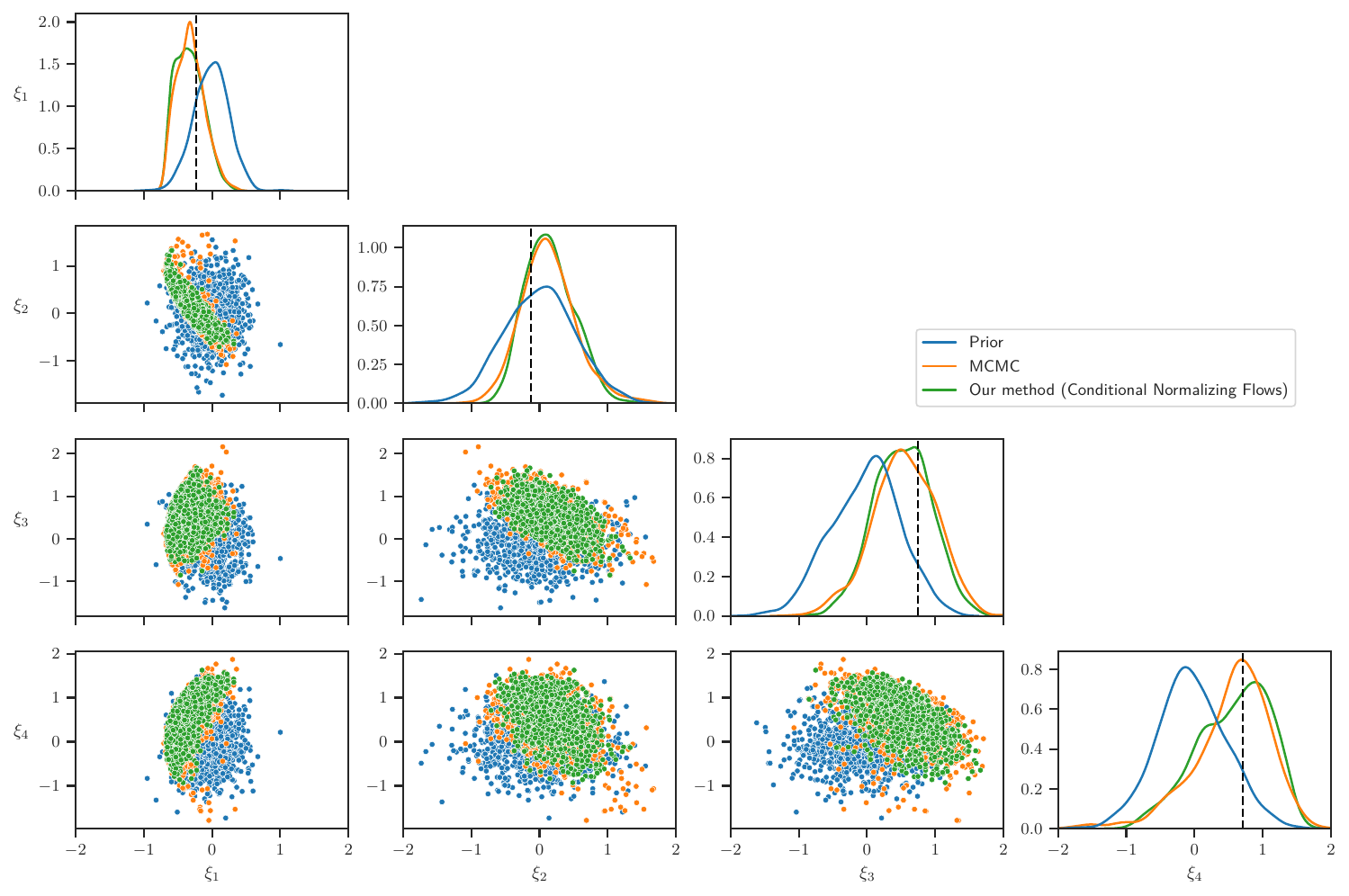}}
\caption{\textcolor{black}{(Inverse Kinematics - Observation set 2.)
Qualitative results of the inverse kinematic problem from Our method (Conditional normalizing flow) and MCMC approaches for the case where the arm ends at a position $y=(1.15, 0.96)$.}}
\label{fig:IK-CNF2}
\end{figure}

\begin{figure}[H]
\centering
\subcaptionbox{Posterior arm configurations from our method.\protect\label{fig:IK-CNF3_AVI}}%
{\includegraphics[width=0.45\textwidth]{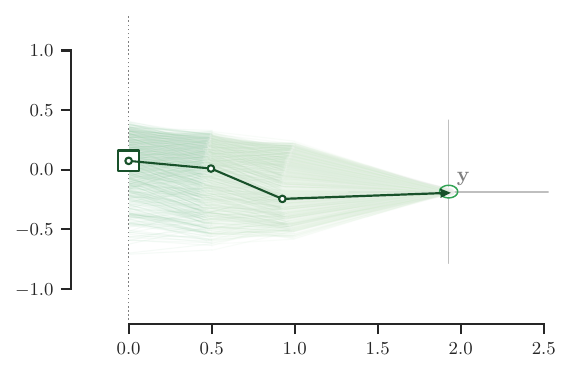}}
\subcaptionbox{Posterior arm configurations from MCMC.\protect\label{fig:IK-CNF3_MCMC}}%
{\includegraphics[width=0.45\textwidth]{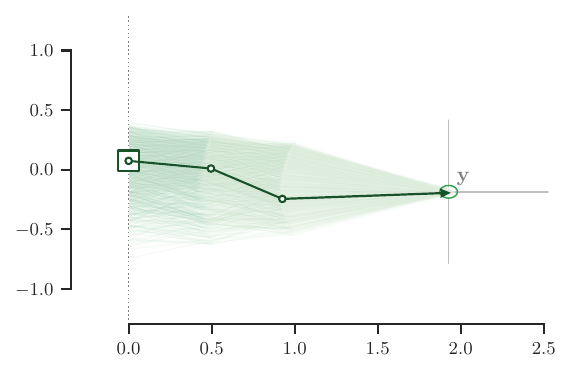}}\\
\subcaptionbox{Pairwise posterior density estimates.\protect\label{fig:IK-CNF3_pairplot}}%
{\includegraphics[width=0.9\textwidth]{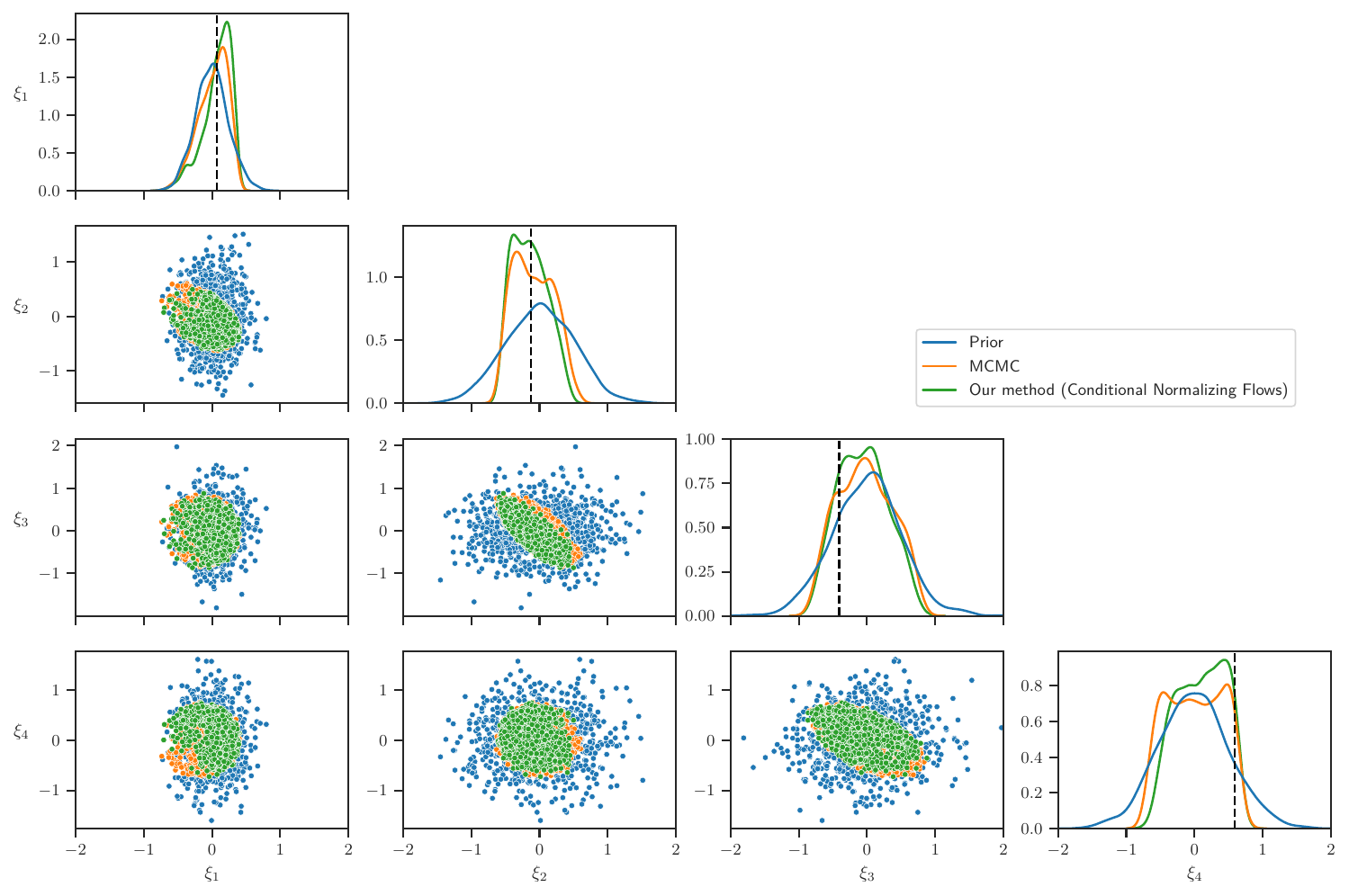}}
\caption{\textcolor{black}{(Inverse Kinematics - Observation set 3.)
Qualitative results of the inverse kinematic problem from Our method (Conditional normalizing flow) and MCMC approaches for the case where the arm ends at a position $y=(1.93, -0.18)$.}}
\label{fig:IK-CNF3}
\end{figure}

\begin{figure}[H]
\centering
\subcaptionbox{Posterior arm configurations from our method.\protect\label{fig:IK-CNF4_AVI}}%
{\includegraphics[width=0.45\textwidth]{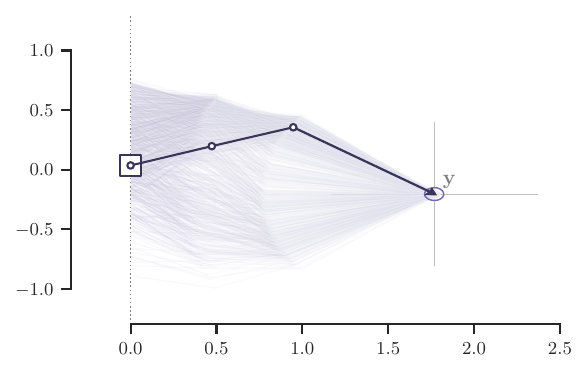}}
\subcaptionbox{Posterior arm configurations from MCMC.\protect\label{fig:IK-CNF4_MCMC}}%
{\includegraphics[width=0.45\textwidth]{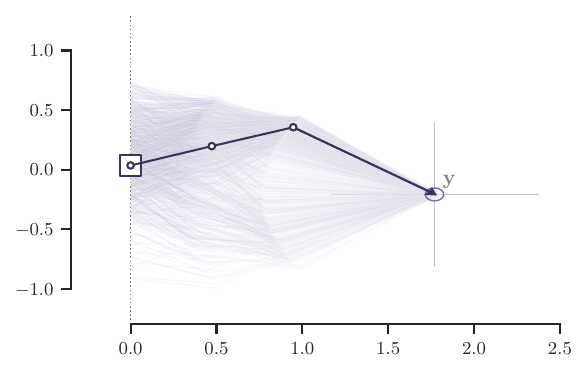}}\\
\subcaptionbox{Pairwise posterior density estimates.\protect\label{fig:IK-CNF4_pairplot}}%
{\includegraphics[width=0.9\textwidth]{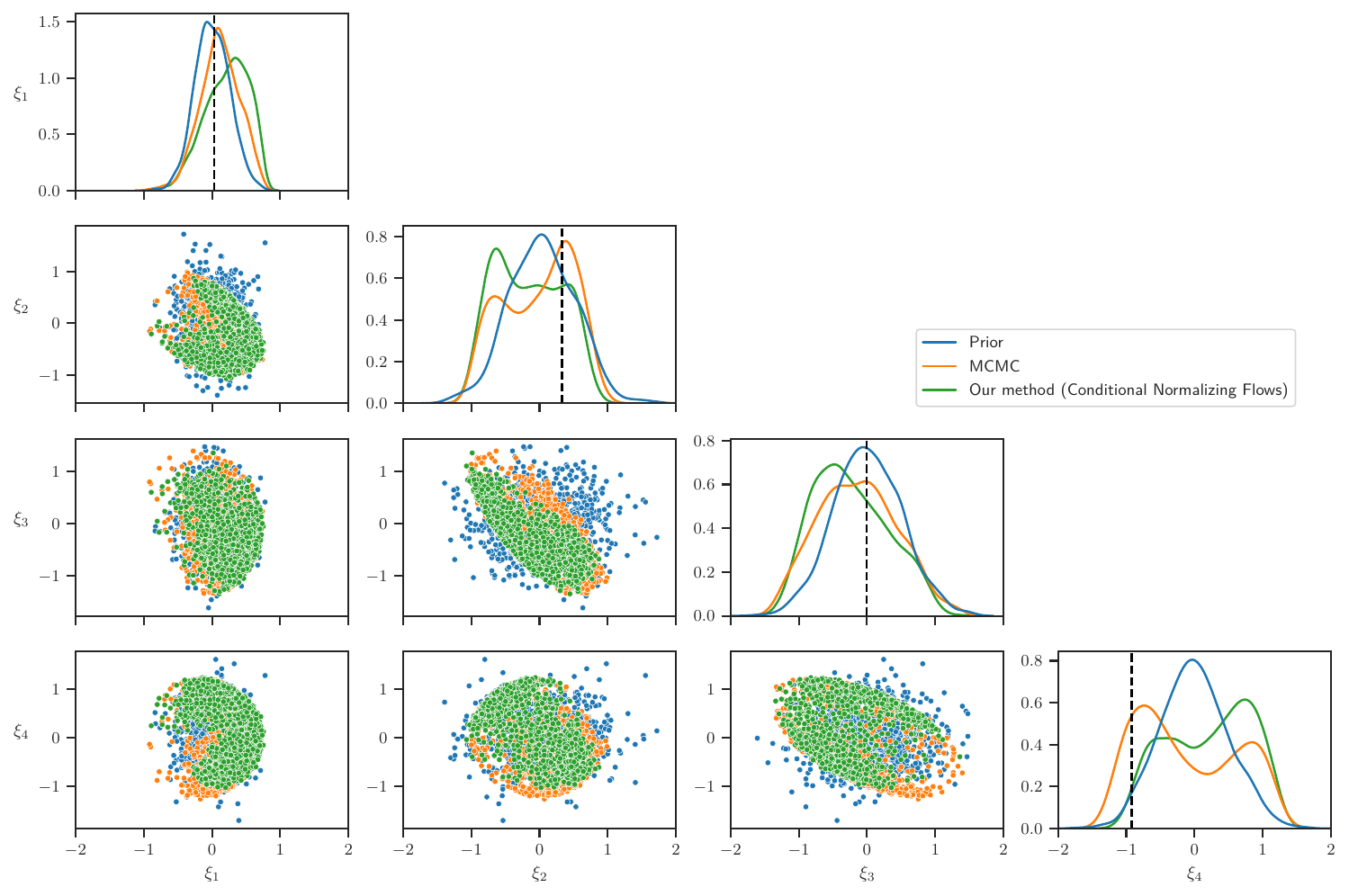}}
\caption{\textcolor{black}{(Inverse Kinematics - Observation set 4.)
Qualitative results of the inverse kinematic problem from Our method (Conditional normalizing flow) and MCMC approaches for the case where the arm ends at a position $y=(1.77, -0.21)$.}}
\label{fig:IK-CNF4}
\end{figure}

\begin{figure}[H]
\centering
\subcaptionbox{Posterior arm configurations from our method.\protect\label{fig:IK-CNF5_AVI}}%
{\includegraphics[width=0.45\textwidth]{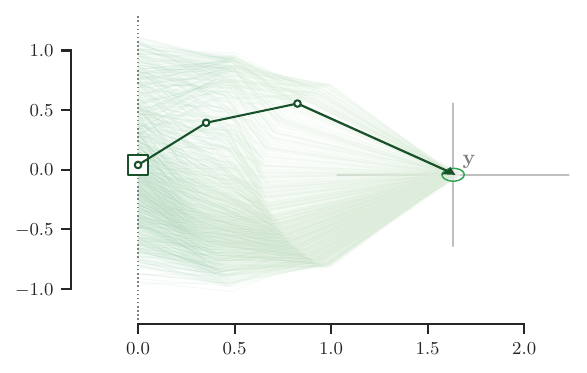}}
\subcaptionbox{Posterior arm configurations from MCMC.\protect\label{fig:IK-CNF5_MCMC}}%
{\includegraphics[width=0.45\textwidth]{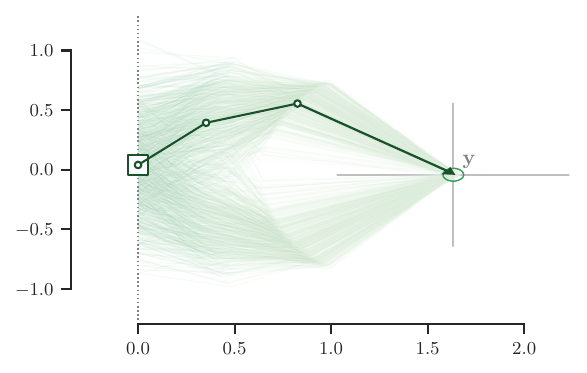}}\\
\subcaptionbox{Pairwise posterior density estimates.\protect\label{fig:IK-CNF5_pairplot}}%
{\includegraphics[width=0.9\textwidth]{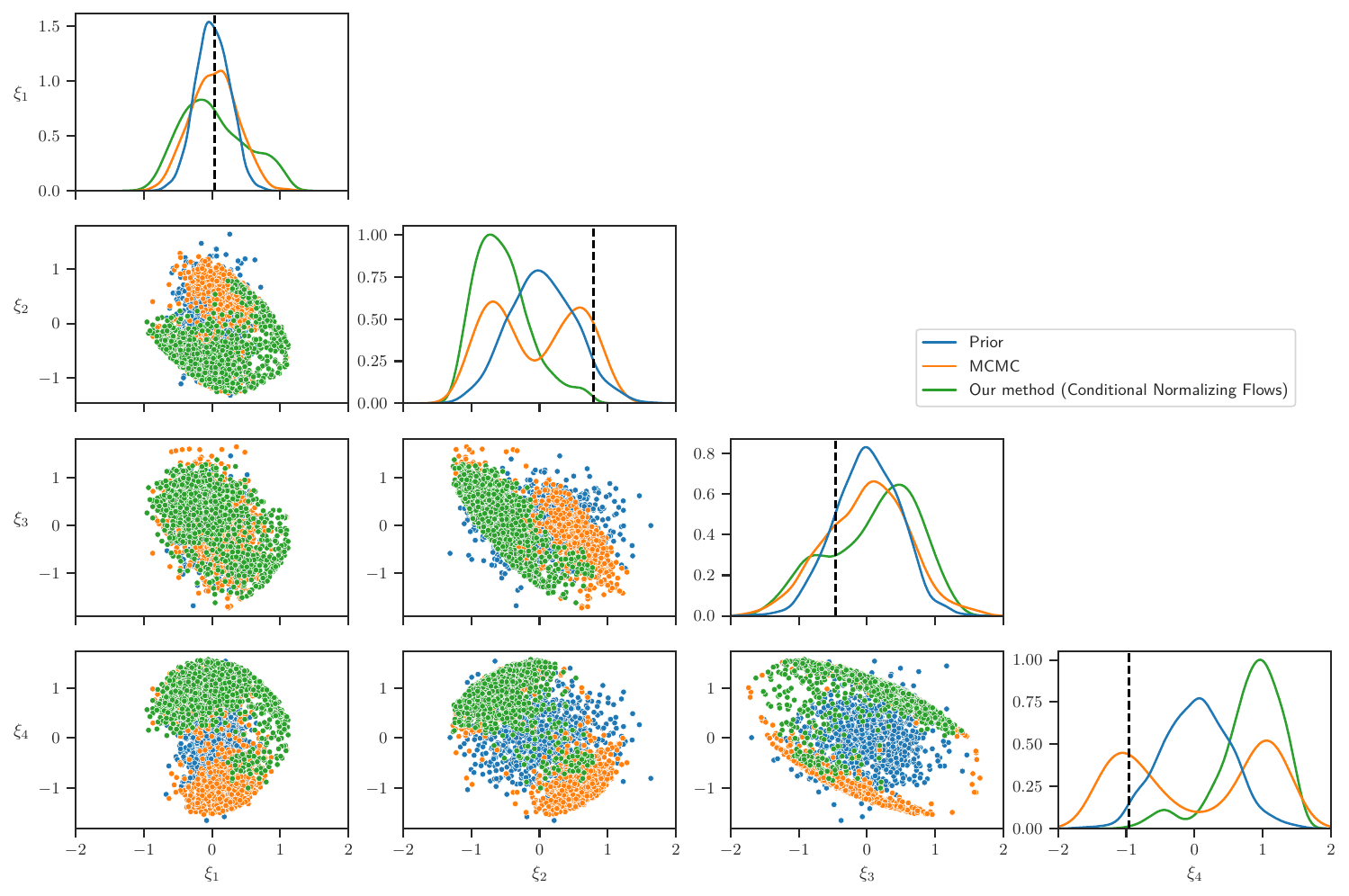}}
\caption{\textcolor{black}{(Inverse Kinematics - Observation set 5.)
Qualitative results of the inverse kinematic problem from Our method (Conditional normalizing flow) and MCMC approaches for the case where the arm ends at a position $y=(1.63, -0.04)$.}}
\label{fig:IK-CNF5}
\end{figure}

\textcolor{black}{Qualitative results of the posteriors learned using this flow model for the same five endpoint observation $y$ cases discussed above are shown in Figs.~(\ref{fig:IK-CNF1}-\ref{fig:IK-CNF5}).
From these, we can clearly see that our conditional flow guide is able to capture all possible arm configurations without any issues.
These results clearly show empirically that the conditional flow models are able to almost learn the complex target posteriors with non-Gaussian nature and multi-modalities in one shot.
However, from the pairplots depicted in Fig.~\ref{fig:IK-CNF4} and Fig.~\ref{fig:IK-CNF5}, we see that the flow model is not able to capture the multiple modes in the true posterior completely.
This is probably happening because the flow model is struck at a local optimal solution.
Additionally, we know that there will be some amortization gap.
This gap ideally should decrease with the increase of flow model length and flow model inner network sizes. 
However, we do not know the ideal flow configuration and length for achieving zero gap.}

\begin{figure}[H]
  \centering
  \subfloat{\includegraphics[width=0.45\textwidth]{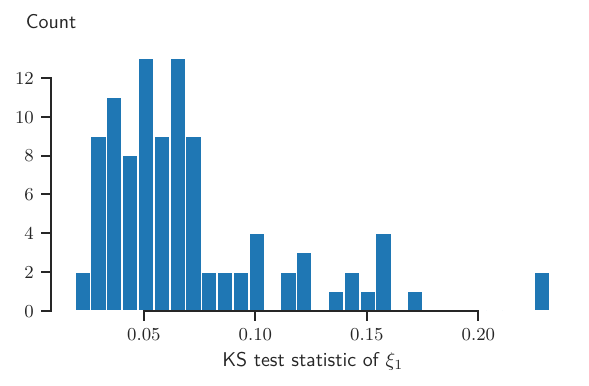}}
  \subfloat{\includegraphics[width=0.45\textwidth]{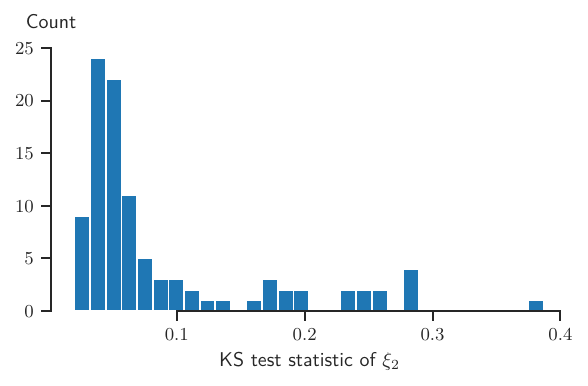}}\\
  \subfloat{\includegraphics[width=0.45\textwidth]{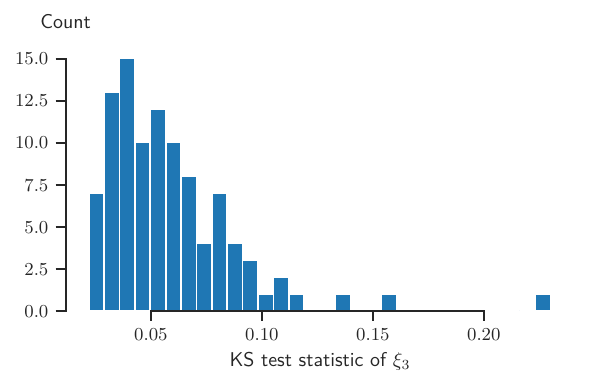}}
  \subfloat{\includegraphics[width=0.45\textwidth]{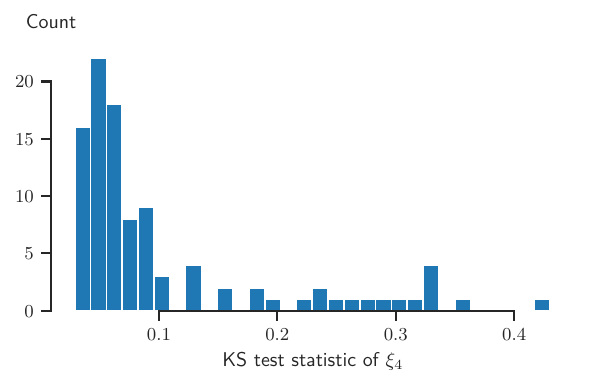}}
  \caption{\textcolor{black}{(Inverse Kinematics) Histograms of KS test statistic values of parameter posteriors with Conditional normalizing flow guide for $N_{y} = 100$ samples from the data density.}}
 \label{fig:IK-CNF_hist}
\end{figure}

\textcolor{black}{Nevertheless, the re-simulation error with using these conditional flow models ($\mathcal{E}_{\text{re-sim}} = \num{1.79e-2}$) is very low compared to re-simulation error with amortized Gaussian guide discussed above.
Even the error metrics based on KS test statistic values are skewed towards the right as in Fig.~\ref{fig:IK-CNF_hist} with very low values.
Overall, these results demonstrate the effectiveness and accuracy of our proposed amortization approach for inferring complex posterior distributions on the fly using conditional flow guides.}

%
%
%
%
%
%
%

%

\newpage
\section{Conclusion}
\label{sec:conclusion}
\noindent

In this work, we developed a methodology for learning Bayesian inverse maps from the observed data to posteriors by using an amortization network.
\textcolor{black}{The amortization network is a black box  
 neural net based model, that takes in observation data as input and outputs the corresponding posterior.}
By using this amortization network, we avoided the need to compute per observation variational parameters and instead, we computed the amortization network parameters which are a set of global variational parameters that generalize over all observations.
We learned these amortization network parameters with an amortized approach for variational inference by taking an additional expectation over standard ELBO with respect to all observations compatible with the model. 
\textcolor{black}{We also proved theoretically that, under certain assumptions, optimizing the proposed amortized loss is equivalent to solving all possible VI problems in one shot.
Towards this end note that, once the amortization network is trained posteriors of observation data are available just at the forward pass of the network thereby enabling real-time on-the-fly inference.}

\textcolor{black}{
In this work, we employed two types of inference models.
In the first inference model, we parametrized the posterior as a distribution belonging to a tractable family, specifically choosing a full-rank Gaussian distribution. Here, the amortization network takes observation data as input and outputs the corresponding posterior parameters: mean vector and covariance matrix.
Now, moving to the second inference model, to parameterize arbitrarily complex posteriors belonging to an intractable family, we used conditional generative models called conditional normalizing flows.
These flow-based models represent complex posterior densities by applying a series of invertible and differentiable transformations based on the data to simple conditional densities. 
Here, the amortization network takes observation data and samples from a simple conditional base distribution as input and outputs the corresponding posterior samples.
Note that the amortization network architecture in the first inference model is simple and straightforward to choose and learn, whereas in the second inference model, choosing the right amortization network architecture is involved and training is time-consuming., nevertheless, we can learn any complex posterior with this model.}
We demonstrated the performance of our amortization networks through three examples.
The posteriors estimated from our amortization network are consistent with the ground truth posteriors from MCMC.


\section{Acknowledgements}
This work has been made possible by the financial support provided by AFOSR program on materials for extreme environments under the grant number FA09950-22-1-0061.
%
%
\bibliography{bibliography.bib}
\end{document}